\documentclass[10pt,twocolumn]{article}
\pdfoutput=1

\usepackage{iccv}
\usepackage{times}
\usepackage{epsfig}
\usepackage{graphicx}
\usepackage{amsmath}
\usepackage{amssymb}
\usepackage[accsupp]{axessibility}

\usepackage{amsmath}
\usepackage{amssymb}
\usepackage{booktabs}
\usepackage{comment}
\usepackage{multirow}
\usepackage{amsthm}
\usepackage{amsmath}
\usepackage{amssymb}
\usepackage{bm}
\usepackage{arydshln}
\usepackage[T1]{fontenc}
\usepackage{cite}
\usepackage{bbold}
\usepackage{mathtools}
\usepackage{dsfont}
\usepackage{algorithm,algcompatible,amssymb,amsmath}
\renewcommand{\COMMENT}[2][.2\linewidth]{%
  \leavevmode\hfill\makebox[#1][l]{//~#2}}
\algnewcommand\algorithmicto{\textbf{to}}
\algnewcommand\RETURN{\State \textbf{return} }

\usepackage{parcolumns}
\usepackage{mdframed}
\usepackage[frozencache,cachedir=.]{minted}

\usepackage[pagebackref=true,breaklinks=true,colorlinks,bookmarks=false]{hyperref}

 \iccvfinalcopy 

\def\iccvPaperID{10718} 

\ificcvfinal\fi

\begin{document}

\title{Adversarial Finetuning with Latent Representation Constraint\\to Mitigate Accuracy-Robustness Tradeoff}

\author{
	Satoshi Suzuki $^{1}$ \quad 
	Shin'ya Yamaguchi  $^{1,2}$ \quad
	Shoichiro Takeda $^{3}$ \quad
	Sekitoshi Kanai $^{1}$ \\
	Naoki Makishima $^{1}$ \quad
	Atsushi Ando $^{1,3}$ \quad
	Ryo Masumura $^{1}$ \vspace{.2em}\\
	$^{1}$NTT Computer and Data Science Laboratories \hspace{0.5cm} $^{2}$Kyoto University\\
        $^{3}$NTT Human Informatics Laboratories 
	\vspace{.4em}\\
	{\tt\small \{satoshixv.suzuki, shinya.yamaguchi, shoichiro.takeda, sekitoshi.kanai,} \\
        {\tt\small naoki.makishima, atsushi.ando, ryo.masumura\}@ntt.com}
}

\maketitle
\ificcvfinal\thispagestyle{empty}\fi

\begin{abstract}
This paper addresses the tradeoff between standard accuracy on clean examples and robustness against adversarial examples in deep neural networks~(DNNs).
Although adversarial training~(AT) improves robustness, it degrades the standard accuracy, thus yielding the tradeoff.
To mitigate this tradeoff, we propose a novel AT method called ARREST, which comprises three components: (i) adversarial finetuning~(AFT), (ii) representation-guided knowledge distillation~(RGKD), and (iii) noisy replay~(NR).
AFT trains a DNN on adversarial examples by initializing its parameters with a DNN that is standardly pretrained on clean examples.
RGKD and NR respectively entail a regularization term and an algorithm to preserve latent representations of clean examples during AFT.
RGKD penalizes the distance between the representations of the standardly pretrained and AFT DNNs.
NR switches input adversarial examples to nonadversarial ones when the representation changes significantly during AFT.
By combining these components, ARREST achieves both high standard accuracy and robustness.
Experimental results demonstrate that ARREST mitigates the tradeoff more effectively than previous AT-based methods do.
\end{abstract}

\section{Introduction}
\label{sect:intro}
Deep neural networks~(DNNs) have demonstrated impressive performance for various computer vision tasks~\cite{Krizhevsky12,Simonyan14c_ICLR,Szegedy14,He16,Redmon17,Long15}.
However, standardly trained DNNs can easily be deceived by adversarial examples~\cite{Szegedy14b,Goodfellow15}, causing incorrect predictions.
Such adversarial examples are images with maliciously designed, human-imperceptible perturbations to deceive a DNN.
As DNNs penetrate almost every corner of our daily life~(\eg, autonomous driving), defense techniques against adversarial examples are becoming increasingly important.

\begin{figure}[t]
  \centering
  \vspace{-2mm}
   \includegraphics[width=59mm]{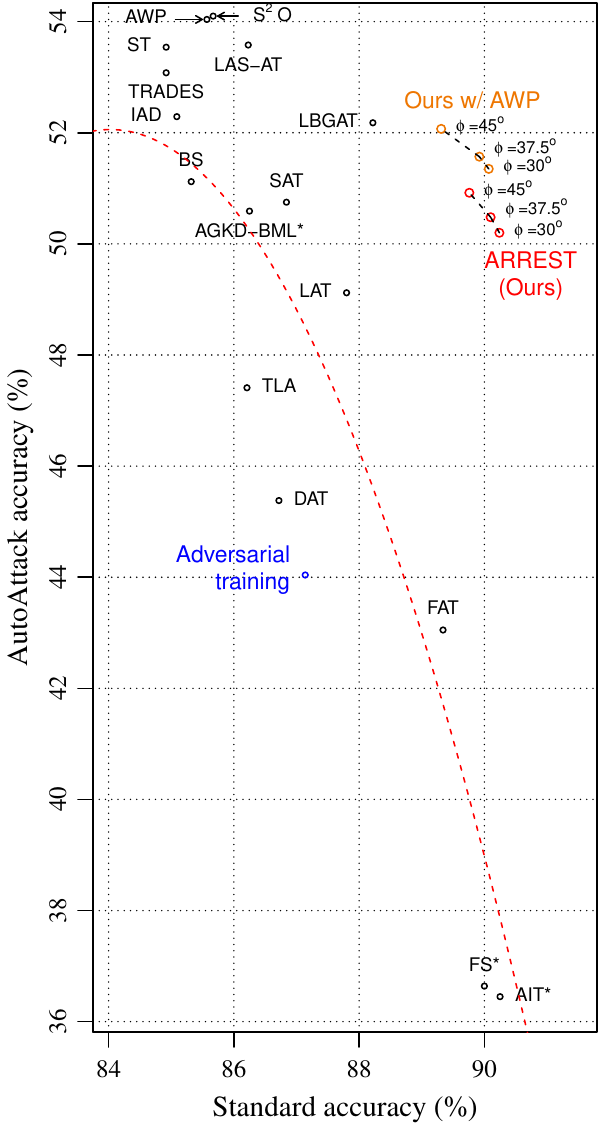}
  \caption{Relationship between the standard and AutoAttack accuracies of various existing methods~(see Appendix~\ref{sec:app_exist}) and our proposed method~(ARREST) on CIFAR-10.
  * indicates a result obtained with WideResNet-28-10~\cite{Zagoruyko16}; the other results were obtained with WideResNet-34-10.
  We also evaluated our method by integrating it with the state-of-the-art AWP method~\cite{Wu20}, as denoted by orange points.
  The red dashed line is an approximated curve of the accuracy-robustness tradeoff.
  }
  \vspace{-4mm}
  \label{fig:intro_res}
\end{figure}

The many defense techniques include feature squeezing~\cite{Xu17}, input denoising~\cite{Guo18,Samangouei18}, adversarial detection~\cite{Ma18,Lee18}, gradient regularization~\cite{Papernot17}, and adversarial training~\cite{Goodfellow15,Madry18}.
Among these, adversarial training~(AT) has attracted much attention as a promising defense method.
AT attempts to build a robust DNN through training on adversarial examples that are generated online to maximally deceive the on-training DNN~\cite{Madry18}.
Since AT's effectiveness was demonstrated by M\k{a}dry~\etal, a remarkable number of improvements have been proposed~\cite{Cai18,Mao19,Zhang19_TRADES,Wang19_BAT,Wang19_DAT,Zhang19_FS,Kumari19,Zhang20_FAT,Zhang20_AIT,Wu20,Kim20,Wang20,Sitawarin21,Wang21,Cui21,Pang21,Chen22,Jia22,Zhu22,Li23,Jin22}.

Although AT is the {\it de facto} standard method to build DNNs that are robust against adversarial examples, it has the disadvantage of degrading the classification accuracy on clean examples~(\ie, the standard accuracy).
This implies a tradeoff between the adversarial robustness and standard accuracy~\cite{Tsipras18,Ilyas19}, which we refer to as the \textit{accuracy-robustness tradeoff}.
Figure~\ref{fig:intro_res} shows the accuracy-robustness tradeoff on CIFAR-10~\cite{Krizhevsky09} for various existing methods evaluated by AutoAttack~\cite{Croce20}.
One particular state-of-the-art method, AWP~\cite{Wu20}, achieves high robustness, but its standard accuracy is 85.57~\%, which is degraded from 95.37~\% with a standardly trained DNN.
This tradeoff limits the practical applications of AT, as many real-world DNN applications require high standard accuracy and cannot afford much degradation.

Several studies~\cite{Zhang19_TRADES,Zhang20_FAT,Cui21,Sitawarin21} have attempted to mitigate the tradeoff; however, the standard accuracy is still degraded from the original accuracy of a standardly trained DNN.
One possible reason for this degradation is the distribution mismatch~\cite{Stutz19,Xie20}, which indicates that clean and adversarial examples have different underlying distributions~\cite{Stutz19,Xie20}.
This mismatch suggests that if we train a robust DNN from scratch, like in the above studies, the latent representation will significantly diverge from that of a standardly trained DNN on clean examples~(see Table~\ref{tbl:cossim}).
Hence, there is room for improvement in terms of obtaining suitable latent representations of both clean and adversarial examples.

In this paper, we propose a novel method to mitigate the accuracy-robustness tradeoff in AT, called AdversaRial finetuning with REpresentation conSTraint~(\textbf{ARREST}).
The idea behind our method is to obtain suitable representations of adversarial examples while preserving suitable representations of clean examples from standardly trained DNNs.
To this end, ARREST comprises three key components: (i)~\textit{adversarial finetuning}~(AFT), (ii)~\textit{representation-guided knowledge distillation}~(RGKD), and (iii)~\textit{noisy replay}~(NR).
ARREST uses a two-step training process for robust DNNs, with standard pretraining of DNNs on clean examples followed by finetuning on adversarial examples to increase robustness.
We especially refer to the second step as AFT.
AFT encourages a DNN to obtain suitable representations of both clean and adversarial examples through finetuning with a standardly pretrained DNN, in contrast to previous studies that trained the DNN from scratch~\cite{Zhang19_TRADES,Zhang20_FAT,Cui21,Sitawarin21}.
We also propose RGKD and NR to preserve representations of clean examples from the pretrained DNN by alleviating the distribution mismatch issue~\cite{Stutz19,Xie20} during AFT.
Inspired by knowledge distillation~\cite{Hinton15_distillation,Romero15}, RGKD penalizes the distance between the on-training DNN's representation and that of the pretrained DNN.
While RGKD modifies the objective function of training, NR modifies the perturbation of inputs in AFT.
When the on-training DNN's representation of a certain clean example significantly diverges from that of the pretrained DNN, NR switches the input from an adversarial example to a noisy one, which is a clean example with added uniform random noise.
NR thus serves to ``remind'' the DNN of the standard pretraining and encourage representations of clean examples to be close to the pretrained DNN's original representations.

We experimentally demonstrate that ARREST achieves an impressive performance.
For example, Fig.~\ref{fig:intro_res} shows its qualitative effectiveness in mitigating the accuracy-robustness tradeoff, as the results for ARREST are clustered on the upper-right side.
Furthermore, we quantitatively evaluate the degree of tradeoff mitigation with a new metric inspired by the BD-Rate~\cite{Bjontegaard01,Strom20} utilized in the field of video compression research.
Specifically, our metric calculates the distance from the tradeoff by approximating a curve to represent it~(red dashed line in Fig.~\ref{fig:intro_res}).
In terms of this metric, ARREST achieves a state-of-the-art performance, thus confirming both its qualitative and quantitative effectiveness.

Our main contributions are threefold:
\begin{enumerate}
    \item We propose a novel adversarial training method, ARREST, to mitigate the accuracy-robustness tradeoff.
    ARREST comprises three components that work complementarily to obtain suitable representations of both clean and adversarial examples.
    \item We conduct a wide range of experiments to demonstrate ARREST's effectiveness.
    Overall, the experimental results provide insights into the strengths of ARREST and the properties of its components.
    \item We propose a novel quantitative evaluation metric inspired by the BD-Rate, and we show that ARREST achieves state-of-the-art performance in terms of this metric.
\end{enumerate}

\section{Related Work}
\label{sect:relate}

\noindent
\textbf{Adversarial Attacks.}
Because of the documented vulnerability of DNNs~\cite{Szegedy14b}, many works have proposed novel adversarial attack techniques~\cite{Goodfellow15,Moosavi-Dezfooli16,Carlini17,Madry18}.
For example, M\k{a}dry~\etal~\cite{Madry18} proposed a projected gradient descent~(PGD) method, which is a multistep version of the fast gradient sign method~(FGSM)~\cite{Goodfellow15}.
Recently, Croce and Hein~\cite{Croce20} proposed two improved versions of the PGD attack, namely APGD-CE and APGD-DLR, which do not require selecting a step size or alternating a loss function, unlike the original PGD.
Then, they combined those two methods with two other complementary adversarial attacks~(FAB~\cite{Croce20_FAB} and Square~\cite{Andriushchenko20}) to evaluate robustness through an approach called AutoAttack.
Recent studies have widely used AutoAttack to evaluate robustness, because it provides more reliable evaluation than the traditional PGD-based evaluation.
Croce and Hein also applied AutoAttack on tens of previous AT-based methods and provided a comprehensive leaderboard~\cite{Croce21}.
In this paper, we mainly apply AutoAttack to evaluate the adversarial robustness, given that it is common and reliable.

\noindent
\textbf{Adversarial Training.}
Many defense methods have been proposed to improve model robustness against adversarial attacks.
Among them, adversarial training~(AT)~\cite{Goodfellow15,Madry18} has attracted much attention.
AT attempts to build robust DNNs through training with online-generated adversarial examples that try to maximally deceive the on-training DNN.
Goodfellow~\etal~\cite{Goodfellow14} used FGSM to generate the adversarial examples; more recently, M\k{a}dry~\etal~\cite{Madry18} used the PGD method.
Because of its high robustness, AT with the PGD method is currently the {\it de facto} standard method to build robust DNNs against adversarial examples.

\noindent
\textbf{Mitigation of Accuracy-Robustness Tradeoff.}
Several studies have attempted to mitigate the accuracy-robustness tradeoff.
Zhang~\etal~\cite{Zhang19_TRADES} proposed a defense method called TRADES, which adjusts the tradeoff with a hyperparameter.
TRADES is based on adversarial logit pairing~(ALP)~\cite{Kannan18}, which increases robustness by encouraging the outputs from clean examples and adversarial examples to be similar to each other.
Cui~\etal~\cite{Cui21} proposed a method that guides a DNN's output to be the same as that of a standardly trained DNN, and they demonstrated that this method mitigates the tradeoff more than TRADES or ALP.
Zhang~\etal~\cite{Zhang20_FAT} and Sitawarin~\etal~\cite{Sitawarin21} attempted to mitigate the tradeoff by using a curriculum learning strategy~\cite{Bengio09}.
Although these are important methods that address the accuracy-robustness tradeoff in AT, they still degrade the standard accuracy from the original accuracy of a standardly trained DNN.
We argue that this degradation is due to the distribution mismatch.
In ARREST, we use three complementary components to address this issue.
As another methodology to address the distribution mismatch, Xie~\etal~\cite{Xie20} proposed using different batch normalization layers~\cite{Ioffe15} for clean and adversarial examples.
However, their approach requires knowing at test time whether an input example is clean or adversarial, which may not be practical.
%

In another line of research, a methodology has been proposed to mitigate the accuracy-robustness tradeoff by using additional real or synthetic examples for training~\cite{Raghunathan20,Rade21,Gowal21}.
In general, AT requires more training data to generalize a DNN than standard training does~\cite{Schmidt18}, and DNNs often suffer from overfitting during AT~\cite{Rice20}.
This methodology can alleviate the overfitting and thus mitigate the accuracy-robustness tradeoff.
However, the use of additional examples often leads to prohibitive increases in the training time cost.
Thus, we focus primarily on methods that do not require additional examples.
Note that we will also demonstrate that using the additional examples in \cite{Raghunathan20} can provide benefits for ARREST~(see Table~\ref{tbl:RST}).

\noindent
\textbf{Adversarial Finetuning.}
AFT has been used in several recent studies~\cite{Jeddi20,Moosavi-Dezfooli19}.
Unlike ARREST, however, the aim of those studies was not mitigation of the accuracy-robustness tradeoff.
Jeddi~\etal~\cite{Jeddi20} used AFT to make AT faster by reducing the number of training epochs.
Moosavi-Dezfooli~\etal~\cite{Moosavi-Dezfooli19} analyzed the effect of AT by comparing a DNN's decision boundary before and after AFT was applied.
In contrast, ARREST aims to apply AFT to mitigate the accuracy-robustness tradeoff by incorporating RGKD and NR.

\begin{figure*}[t]
  \centering
   \includegraphics[width=175mm]{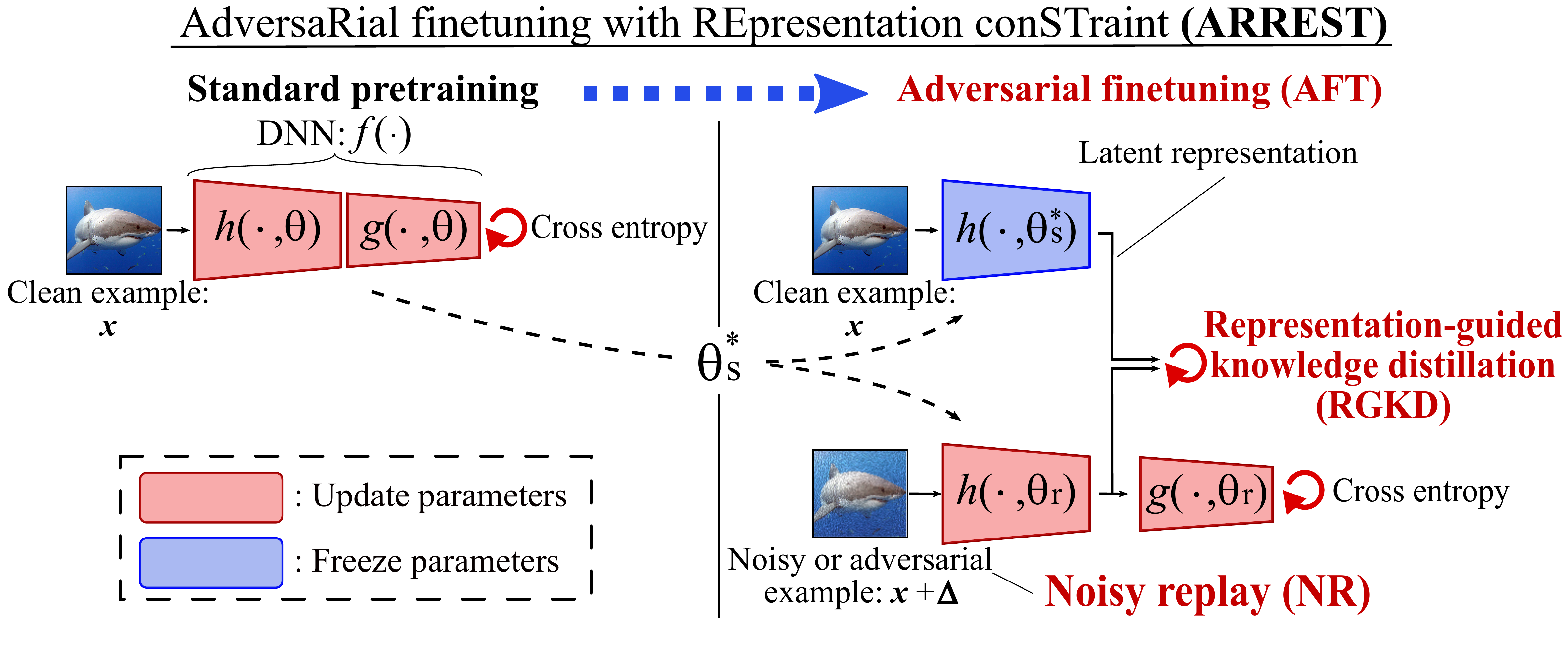}
  \caption{Overview of the proposed method, ARREST.}
  \label{fig:prop}
\end{figure*}

\section{Preliminaries}
\label{sect:prel}

We first describe the conventional AT method~\cite{Madry18}.
In general, AT directly incorporates adversarial examples into the training process to solve the following min-max problem:
\begin{align}
\centering
    \underset{\theta_{\rm r}}{{\rm min}}~\underset{(\bm{x},y) \sim \mathds{D}}{\mathds{E}}
    \left[\underset{||\bm{\delta}||_p \leq \varepsilon}{{\rm max}}~\mathcal{L}_{\rm CE} (f(\bm{x} + \bm{\delta}; \theta_{\rm r}), y) \right].
\label{formula:1}
\end{align}
Here, $\theta_{\rm r}$ represents the parameters of the DNN $f(\cdot)$, and $\mathcal{L}_{\rm CE} (\cdot)$ is the cross-entropy loss, which is commonly used for classification tasks.
$\bm{x}$ and $y$ are a clean training example and its ground-truth label, respectively, which are sampled from an underlying data distribution $\mathds{D}$.
In AT, the cross-entropy loss is calculated with an adversarial example, $\bm{x} + \bm{\delta}$.
The $L_p$-norm of $\bm{\delta}$, $||\bm{\delta}||_p$, is bounded by a perturbation budget $\varepsilon$.
As the inner maximization problem in Eq.~(\ref{formula:1}) cannot be solved in closed form, the PGD method~\cite{Madry18} is commonly used to solve it heuristically.

In the conventional AT method~\cite{Madry18}, a robust DNN is trained from scratch, \ie, the parameters $\theta_{\rm r}$ of $f(\cdot)$ are randomly initialized.
This training style has been followed by almost all AT improvements~\cite{Cai18,Mao19,Zhang19_TRADES,Wang19_BAT,Wang19_DAT,Zhang19_FS,Zhang20_FAT,Zhang20_AIT,Wu20,Kim20,Wang20,Sitawarin21,Wang21,Cui21,Pang21,Chen22,Jia22,Zhu22,Li23,Jin22}.
We found that a robust DNN trained from scratch on adversarial examples obtains significantly different representations from a standardly trained DNN because of the distribution mismatch issue~\cite{Stutz19,Xie20}~(see Table~\ref{tbl:cossim}).
Therefore, such robust DNNs have difficulty obtaining suitable representations of clean examples.

In contrast with the above studies, in this paper, we propose ARREST, which finetunes a standardly pretrained DNN with adversarial examples to increase its robustness, while introducing a constraint on the latent representation.
As a result, ARREST generates a DNN that is robust against adversarial examples but achieves high standard accuracy.
To formulate the constraint, we divide the DNN $f(\cdot)$ into $g \circ h(\cdot)$, where $h(\cdot)$ maps an input example to its corresponding latent representation, and $g(\cdot)$ is a classifier in $f(\cdot)$.
The dividing point between $h(\cdot)$ and $g(\cdot)$ depends on the DNN architecture, though the penultimate layer is usually used~\cite{Mao19,Wang21}.

\section{Proposed Method}

In this paper, we propose ARREST to mitigate the accuracy-robustness tradeoff.
Our main idea is to obtain suitable representations of adversarial examples while preserving suitable representations of clean examples from standardly trained DNNs.
To this end, ARREST comprises three key components: (i)~\textit{adversarial finetuning}~(AFT), (ii)~\textit{representation-guided knowledge distillation}~(RGKD), and (iii)~\textit{noisy replay}~(NR).
Figure~\ref{fig:prop} and Algorithm~\ref{alg1} provide an overview and the detailed procedure of ARREST, respectively.
In this section, we explain each component in detail.

\subsection{Adversarial Finetuning~(AFT)}

We first explain AFT.
In ARREST, we use a two-step training process to obtain robust DNNs, where the steps are standard pretraining of DNNs on clean examples and finetuning of the pretrained DNNs on adversarial examples via RGKD and NR.
We especially refer to the second step as AFT.

Before AFT, we standardly train on clean examples to obtain the following DNN:
\begin{align}
\centering
    \theta_{\rm s}^* = \underset{\theta}{{\rm argmin}}~\underset{(\bm{x},y) \sim \mathds{D}}{\mathds{E}}
    \left[\mathcal{L}_{\rm CE} (f(\bm{x}; \theta), y) \right].
\label{formula:pretrain}
\end{align}
In AFT, we finetune the pretrained DNN via the min-max problem in Eq.~(\ref{formula:1}).
Specifically, as given in Algorithm~\ref{alg1}, we initialize the parameters of a DNN $\theta_{\rm r}$ with those of $\theta_{\rm s}^*$~(line 1) and optimize $\theta_{\rm r}$ iteratively~(line 9).

Through the use of $\theta_{\rm s}^*$ for the initial parameters, AFT helps the DNN obtain suitable representations of clean examples.
However, AFT does not explicitly impose constraints on the DNN's representations.
As a result, the distribution mismatch issue~\cite{Stutz19,Xie20} causes the DNN's representations of clean examples to gradually diverge from the original representations by the standardly pretrained DNN during AFT.
To address this issue, we propose the application of RGKD and NR to constrain the DNN's representations.

\begin{algorithm}[t]
\renewcommand{\algorithmicrequire}{\textbf{Input:}}
\renewcommand{\algorithmicensure}{\textbf{Output:}}                 
\caption{
AdversaRial finetuning with REpresentation conSTraint~(ARREST).
}
\label{alg1}
\begin{algorithmic}[1]
\REQUIRE parameters of standardly trained DNN $\theta_{\rm s}^*$, perturbation budget $\varepsilon$, distance threshold $\tau$, training dataset $\mathds{D}$, learning rate $\eta$, batch size $m$
\ENSURE parameters of DNN $\theta_{\rm r}$
\STATE Initialize parameters $\theta_{\rm r} \leftarrow \theta_{\rm s}^*$ \COMMENT{\textbf{AFT}}
\WHILE {until convergence}
\STATE Sample mini-batch $\{(\bm{x}_i,y_i)\}_{i=1}^m$ from $\mathds{D}$
\FOR{$i = 1,\cdots, m$}
\STATE Calculate $a = d\left(h(\bm{x}_i; \theta_{\rm r}),h(\bm{x}_i; \theta_{\rm s}^*)\right)$
\STATE{Obtain $\bm{\Delta}_i$ \COMMENT{\textbf{NR}}
\begin{align}
\text{~~~~~~~where }
\begin{cases}
\displaystyle \bm{\Delta}_i = \bm{\delta}_i~\text{obtained by PGD} & (a \leq \tau) \\
\displaystyle \bm{\Delta}_i \sim \mathcal{U}(-\varepsilon,\varepsilon) & (a > \tau)
\nonumber
\end{cases}
\end{align}
}
\STATE Calculate $\mathcal{L}(\bm{x}_i, \bm{\Delta}_i, y_i, \theta_{\rm r})$ in Eq.~(\ref{formula:3}) \COMMENT{\textbf{RGKD}}
\ENDFOR
\STATE $\theta_{\rm r} \leftarrow \theta_{\rm r} - \eta \ \frac{1}{m} \sum^m_{i=1} \nabla_{\theta_{\rm r}} \mathcal{L}(\bm{x}_i, \bm{\Delta}_i, y_i, \theta_{\rm r})$
\ENDWHILE
\end{algorithmic}
\end{algorithm}

\subsection{Representation-Guided Knowledge Distillation~(RGKD)}
\label{subsetc:rgkd}
In RGKD, we penalize the distance between the representations of the DNN $\theta_{\rm r}$ and the standardly pretrained DNN $\theta_{\rm s}^*$ during AFT.
RGKD was inspired by the knowledge distillation methodology~\cite{Hinton15_distillation,Romero15} in model compression.
Knowledge distillation was originally proposed to guide a small DNN using knowledge~(\ie, an output or representation) from a large DNN to reduce the computation cost.
Here, we apply this concept to mitigate the accuracy-robustness tradeoff by guiding the DNN with a representation from the standardly pretrained DNN.

We define the loss of RGKD as follows:
\begin{align}
\centering
    \mathcal{L}_{\rm RGKD} (\bm{x}, \bm{\delta}, \theta_{\rm r})
    = d\left(h(\bm{x} + \bm{\delta}; \theta_{\rm r}),h(\bm{x}; \theta_{\rm s}^*)\right),
\label{formula:2}
\end{align}
where $d(\cdot)$ is a distance function, such as the angular distance~\cite{Wang17}, to measure the similarity between two representations.
From the definition of $h(\cdot)$ in Section~\ref{sect:prel}, the arguments of $d(\cdot)$ are the representation by the DNN $\theta_{\rm r}$ of an adversarial example $\bm{x} + \bm{\delta}$ and that by the standardly pretrained DNN $\theta_{\rm s}^*$ of a clean example $\bm{x}$.
Note that the parameters of $\theta_{\rm s}^*$ are not updated~(frozen), as shown in Fig.~\ref{fig:prop}.
By minimizing this loss, we can penalize the DNN's representation if it diverges from the original representation $h(\bm{x}; \theta_{\rm s}^*)$.

In ARREST, the loss for optimizing DNNs is the summation of $\mathcal{L}_{\rm CE}$ and $\mathcal{L}_{\rm RGKD}$:
\begin{align}
\centering
    \mathcal{L}(\bm{x} , \bm{\delta}, y, \theta_{\rm r}) = 
    \mathcal{L}_{\rm CE} (f(\bm{x} &+ \bm{\delta}; \theta_{\rm r}), y) \nonumber \\
    &+ \lambda \ \mathcal{L}_{\rm RGKD} (\bm{x}, \bm{\delta}, \theta_{\rm r}),
\label{formula:3}
\end{align}
where $\lambda$ is a hyperparameter for determining the effect of RGKD on optimization.
As seen in lines 7 and 9 of Algorithm~\ref{alg1}, this loss is calculated across all examples in a mini-batch and used for optimization of $\theta_{\rm r}$.
%

Several AT methods~\cite{Cui21,Wang21} have also used the knowledge distillation methodology and guided a DNN by using a logit~(output)~\cite{Cui21} or a representation transferred to an attention map~\cite{Wang21}.
We found experimentally that RGKD is the best of those methods for mitigating the tradeoff~(see Table~\ref{tbl:comp_const}).

\subsection{Noisy Replay~(NR)}
\label{subsect:nr}

While RGKD modifies the objective function for training, NR addresses the distribution mismatch issue by modifying the perturbation of inputs in AFT.
Specifically, it monitors the distance between $h(\bm{x}; \theta_{\rm r})$ and $h(\bm{x}; \theta_{\rm s}^*)$, \ie, $d\left(h(\bm{x}; \theta_{\rm r}),h(\bm{x}; \theta_{\rm s}^*)\right)$.
When this distance exceeds a predefined threshold, NR attempts to avoid increasing the distance further by utilizing the replay technique~\cite{Farquhar18,Rolnick19,Boschini22}.
This technique was originally developed to address catastrophic forgetting during continual learning~\cite{Mccloskey89}.
It retrains a DNN with data from a previous task during current task training.
Recent research~\cite{Ramasesh21} found that, during a current task, the replay technique preserves a suitable latent representation obtained by the previous task.
In our case, the previous and current tasks correspond to standard pretraining on clean examples and AFT on adversarial examples, respectively.
Via this analogy, NR switches input examples by using $d\left(h(\bm{x}; \theta_{\rm r}),h(\bm{x}; \theta_{\rm s}^*)\right)$:
\begin{equation}
\centering
    \bm{x} + \bm{\Delta},~
\small{\text{where }}
\begin{cases}
 	\displaystyle \small{\bm{\Delta} \sim \mathcal{U}(-\varepsilon,\varepsilon)} & \small{\text{if $d\left(h(\bm{x}; \theta_{\rm r}),h(\bm{x}; \theta_{\rm s}^*)\right) > \tau$}} \\
        \displaystyle \small{\bm{\Delta} = \bm{\delta}} & \small{\text{otherwise}},
\end{cases}
\label{formula:nr}
\end{equation}
where $\mathcal{U}(-\varepsilon,\varepsilon)$ denotes a uniform distribution bounded by the absolute value of $\varepsilon$.
Empirically, we found that adjusting the threshold value $\tau$ plays a role in balancing the tradeoff.
Specifically, the robustness increases as $\tau$ increases, while the accuracy increases as $\tau$ decreases.
Therefore, in the experimental section, we use several $\tau$ values for achieving various tradeoffs~(see Figs.~\ref{fig:intro_res} and \ref{fig:res_100}).

With Eq.~(\ref{formula:nr}), NR inputs random noisy examples to the DNN when the distance exceeds $\tau$.
We could also naively consider inputting a clean example~($\bm{\Delta}=\bm{0}$) when $d\left(h(\bm{x}; \theta_{\rm r}),h(\bm{x}; \theta_{\rm s}^*)\right)$ $>\tau$.
However, we observed that this approach does not work well, and we discuss the reason in Section~\ref{sect:anal}.
The NR processes are in lines 5 and 6 of Algorithm~\ref{alg1}.
Finally, the loss in Eq.~(\ref{formula:3}) is calculated with $\bm{\Delta}$, rather than $\bm{\delta}$; that is, $\mathcal{L}(\bm{x}, \bm{\Delta}, y, \theta_{\rm r})$ is used for optimization in line 7.

\section{Experiments}
\label{sec:exp}

\begin{table*}[tb]
\centering
\caption{Quantitative evaluation of ARREST and four existing state-of-the-art methods via the Sum and ARDist metrics for CIFAR-10 and CIFAR-100 datasets.
        The adversarial robustness was calculated utilizing AutoAttack.
		Bold type indicates the highest value for each metric.}
\label{tbl:sota_c10}
\scalebox{1.0}{
\begin{tabular}{l|cc|cc|cc|cccc}\hline
       & \multicolumn{4}{c|}{CIFAR-10}          & \multicolumn{4}{c}{CIFAR-100}         \\
                    & Standard  & AutoAttack  & Sum    & ARDist & Standard  & AutoAttack  & Sum   & ARDist \\ \hline \hline
AT~\cite{Madry18}   & 87.14\%   & 44.04\%     & 131.18 & -1.500 & 59.59\%   & 22.86\%     & 82.45 & -3.268  \\ 
LAS-AT~\cite{Jia22} & 86.23\%   & 53.58\%     & 139.81 &  2.236 & 61.80\%   & 29.03\%     & 90.83 & 3.189   \\
AWP~\cite{Wu20}     & 85.57\%   & 54.04\%     & 139.61 &  2.314 & 60.38\%   & 28.86\%     & 89.24 & 2.424  \\
S$^2$O~\cite{Jin22} & 85.67\%   & 54.10\%     & 139.77 &  2.410 & 63.40\%   & 27.60\%     & 91.00 & 2.786  \\
LBGAT~\cite{Cui21}  & 88.22\%   & 52.18\%     & 140.40 &  2.706 & 70.25\%   & 26.73\%     & 96.98 & 6.639  \\
ARREST~(ours)       & 90.24\%   & 50.20\%     &\textbf{140.44} & \textbf{3.521}   & 73.05\%  & 24.32\%     & \textbf{97.37} & \textbf{7.165}  \\\hline
  		\end{tabular}
    }
\end{table*}

\subsection{Quantitative Evaluation Metrics}
\label{subsect:metrics}
Before describing our experiments, we introduce a new metric, accuracy robustness distance~(ARDist), to quantitatively evaluate mitigation of the accuracy-robustness tradeoff.
ARDist was inspired by the BD-Rate metric that is commonly used in the field of video compression~\cite{Bjontegaard01,Strom20} and has a similar purpose.
Specifically, the BD-Rate quantitatively evaluates the \textit{tradeoff between a codec's bitrate and distortion} by approximating a curve representing the tradeoff.
Similarly, ARDist uses existing methods to approximate a curve representing the accuracy-robustness tradeoff.
This approximation is easily implemented by polynomial regression with a cubic function on all data points of the existing methods~(listed in Appendix~\ref{sec:app_exist}), including ARREST\footnote{We used the result with $\phi=30^\circ$ for approximation.}.
In this paper, we made this approximation for CIFAR-10 and CIFAR-100 datasets.
The obtained equations of the approximated curves are
%
\begin{align}
\centering
    c_{10}(x) &= (9.877\cdot10^{-5})\ x^3 - 0.3922 x^2 + 63.82 x - 2600, \nonumber \\
    c_{100}(x) &= (5.615\cdot10^{-4})\ x^3 - 0.1582 x^2 + 12.44 x - 271.8, \nonumber
\end{align}
where $x$ indicates the standard accuracy, and $c_{10}(\cdot)$ and $c_{100}(\cdot)$ denote the approximated curves for CIFAR-10 and CIFAR-100, respectively.
The red dashed line in Fig.~\ref{fig:intro_res} is a concrete example of an approximated curve for CIFAR-10, and it fits the accuracy-robustness tradeoffs of the existing methods.
ARDist evaluates the mitigation by calculating the distance between the approximated curve and a point given by the method being evaluated.
This calculation can be done numerically, and we provide Python source code in Appendix~\ref{sec:app_res-func}.
ARDist yields positive or negative values depending on whether a method's point is above or below the approximated curve.

The simple sum of the accuracy and robustness is sometimes used as another quantitative metric~\cite{Sitawarin21}.
This is an important metric for evaluating a method's absolute performance in terms of its accuracy and robustness.
In contrast, ARDist can evaluate the relative performance when comparing the tradeoffs of a new method and existing methods.
In this paper, we use both metrics to quantitatively evaluate the tradeoff mitigation from multiple perspectives.

\subsection{Experimental Settings}
\label{subsect:exp_set}

\noindent
\textbf{Datasets.}
We evaluated ARREST on two popular datasets: CIFAR-10 and CIFAR-100~\cite{Krizhevsky09}.
CIFAR-10 dataset contains 60,000 color images having a size of 32$\times$32 in 10 classes, with 50,000 training and 10,000 test images.
CIFAR-100 dataset contains 50,000 training and 10,000 test images in 100 classes.

\noindent
\textbf{Optimization Details.}
We adopted the SGD optimizer with a momentum of 0.9 and weight decay of $5\times 10^{-4}$.
The batch size was set to 128.
In the standard pretraining, we set the number of training epochs to 100.
The learning rate started at 0.1 and then decayed by $\times 0.1$ with transition epochs $\{75, 90\}$, following Zhang~\etal~\cite{Zhang19_TRADES}.
In AFT, we set the number of training epochs to 20.
The learning rate started at 0.025, decayed to 0.02 at 11 epochs, and then decayed by half every two epochs thereafter.
We used NR only in the first 10 of 20 AFT epochs, which minimized the sacrificed robustness.

\noindent
\textbf{Implementation Details.}
We used WideResNet-34-10~\cite{Zagoruyko16} as the main DNN architecture, following many previous studies~\cite{Madry18,Zhang19_TRADES,Kumari19,Zhang20_FAT,Wu20,Kim20,Wang20,Cui21}.
On CIFAR-10, we also used ResNet-18~\cite{He16} to evaluate ARREST's flexibility with respect to the architecture.
%
Following previous works~\cite{Madry18,Zhang19_TRADES,Cui21}, $\bm{\delta}$ was bounded by the $L_\infty$-norm.
We used a perturbation budget of $\varepsilon=8/255$ for both training and evaluation.
In AFT, $\bm{\delta}$ was obtained by the PGD method with a step size of $2/255$ and 10 iterative steps.
The PGD objective function was $\mathcal{L}_{\rm CE}$ alone, without the loss of RGKD.
For ARREST, we used the output of the penultimate layer~(just before global pooling) as the latent representation for both architectures, as a higher-dimensional penultimate layer tends to preserve more information~\cite{Mao19}.
The distance function was the angular distance~\cite{Wang17}: $d(\bm{u},\bm{v}) = 1 -\frac{|\bm{u} \cdot \bm{v}|}{||\bm{u}||_2\cdot||\bm{v}||_2}$.
Note that RGKD also performs well with other distance functions, \eg, the mean squared error, and we provide those results in Appendix~\ref{sec:app_res_function}.
We set the hyperparameter $\lambda$ to 50.
We adjusted $\tau$ using the form $1 - \cos\phi$.
Since $\tau$ determines the balance of the tradeoff~(as mentioned in Subsection~\ref{subsect:nr}), we set various values for $\phi$: $\{30^\circ, 37.5^\circ, 45^\circ\}$ for CIFAR-10 and $\{30^\circ, 32.5^\circ, 35^\circ\}$ for CIFAR-100.

\begin{figure}[bt]
  \centering
   \includegraphics[width=71.5mm]{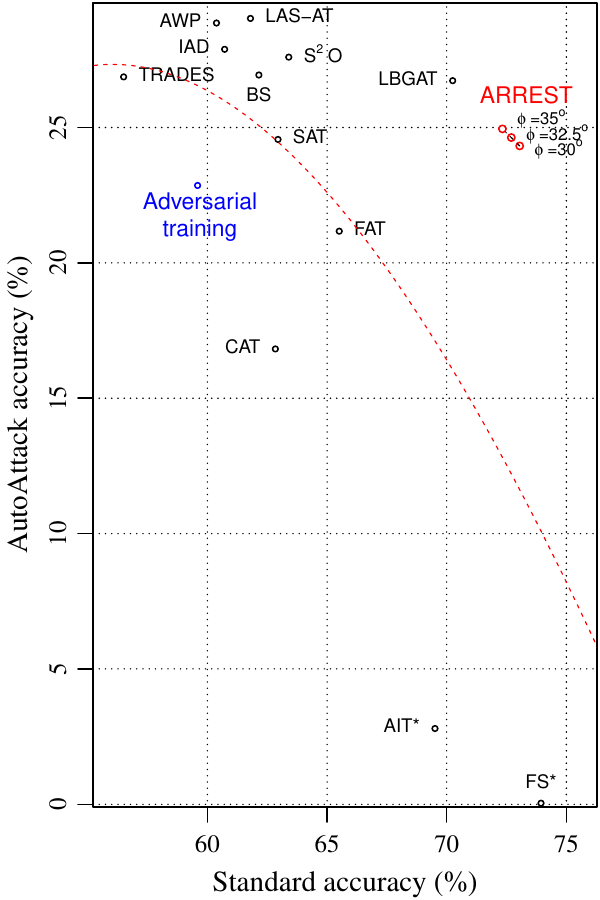}
   \vspace{0.5mm}
  \caption{Relationship between the standard and AutoAttack accuracies of existing methods~(see Appendix~\ref{sec:app_exist}) and ARREST on CIFAR-100.
  * indicates a result obtained with WideResNet-28-10; the other results were obtained with WideResNet-34-10.
  The red dashed line is the tradeoff's approximated curve.}
  \label{fig:res_100}
\end{figure}

\subsection{Comparison with Existing Methods}

To benchmark ARREST's effectiveness, we conducted comparison experiments with existing AT methods on CIFAR-10 and CIFAR-100.
Figures~\ref{fig:intro_res} and \ref{fig:res_100} show the standard accuracy and robustness calculated by AutoAttack~\cite{Croce20} for the existing methods and ARREST on CIFAR-10 and CIFAR-100, respectively.
The details of the existing methods are given in Appendix~\ref{sec:app_exist}.
In both figures, ARREST appears in the upper-right relative to the existing methods.
As this position indicates both high accuracy and high robustness, these results qualitatively demonstrate the effectiveness of ARREST for mitigating the accuracy-robustness tradeoff as compared to existing methods.
It is worth mentioning that ARREST integrates easily with other existing techniques for increasing the robustness, such as AWP~\cite{Wu20}.
In Fig.~\ref{fig:intro_res}, the results of ARREST with AWP~\cite{Wu20} are denoted by the orange points.
We can observe that this combination achieves higher robustness with only a slight sacrifice to the standard accuracy.
As a result, ARREST with AWP better mitigates the accuracy-robustness tradeoff compared with only using ARREST.
In Appendix~\ref{sec:app_trades_hypara}, we further compared ARREST with the results of varying the hyperparameter $\beta$ for TRADES and S$^2$O~(integrated with TRADES).
From these comparisons, we can see that the results of ARREST appear on the right side relative to those of TRADES and S$^2$O.
This indicates that ARREST can achieve higher standard accuracy while achieving the same robustness as these methods.

In addition to this qualitative comparison, we conducted a quantitative comparison using the two metrics explained in Subsection~\ref{subsect:metrics}, \ie, ARDist and the sum of the standard accuracy and robustness~(Sum).
Here, we set $\phi$ in NR to $30^\circ$ for both CIFAR-10 and CIFAR-100.
Table~\ref{tbl:sota_c10} lists these results, which were obtained by comparing ARREST with AT~\cite{Madry18} and four existing state-of-the-art methods~(AWP~\cite{Wu20}, LAS-AT~\cite{Jia22}, S$^2$O~\cite{Jin22}, and LBGAT~\cite{Cui21}).
Note that these four existing methods had the best scores in terms of Sum and ARDist across all methods listed in Appendix~\ref{sec:app_exist} on both CIFAR-10 and CIFAR-100.
As shown in the table, ARREST also achieved a state-of-the-art performance in terms of Sum and ARDist on both datasets.
These results indicate that ARREST can obtain more suitable latent representations of both clean and adversarial examples compared to the existing methods.

\subsection{Ablation Study}
\label{subsect:ablation}

We analyzed the ablation effect of each component of ARREST on CIFAR-10 dataset, where ResNet-18 and WideResNet-34-10 were used as the architectures.
Table~\ref{tbl:result} lists the results obtained with four variations of our method~(only AFT, AFT with RGKD, AFT with NR, and all components, \ie, ARREST) and two baselines~(standard training and AT~\cite{Madry18}).
For AFT with NR only, we set $\phi = 45^\circ$ because the constraining effect changes in the absence of RGKD, and $\phi = 30^\circ$ does not yield an optimal performance.
This $\phi$ value was searched from $15^\circ$ to $75^\circ$ in $15^\circ$ increments.
We used AutoAttack~\cite{Croce20} for the evaluation.
The results using other attacks are listed in Appendix~\ref{sec:app_other_attack}.

\begin{table}[tb]
	\centering
		\caption{Results obtained with four variations of our method~(only AFT, AFT with RGKD, AFT with NR, ARREST) and two baselines~(standard training and AT).
        }
        \label{tbl:result}
		\scalebox{0.83}{
  		\begin{tabular}{l|cc|cccccccccc}  \hline
              & \multicolumn{2}{c}{ResNet-18} & \multicolumn{2}{|c}{WideResNet-34-10} \\
                    & Standard         & AutoAttack     & Standard & AutoAttack \\ \hline\hline
	   ST        & \textbf{94.53}\%  & 0\%    & \textbf{95.37}\% & 0\% \\
AT~\cite{Madry18}   & 84.71\% & 44.19\%  & 87.14\% & 44.04\%\\ \hdashline
AFT                 & 84.08\% & 45.36\%  & 87.54\% & 48.74\% \\
          AFT + RGKD& 85.11\% & 46.01\%  & 88.52\% & \textbf{50.20}\%\\ 
	   AFT + NR  & 85.52\% & 45.29\%  & 88.94\% & 48.63\% \\
	   ARREST    & 86.63\% & \textbf{46.14}\% & 90.24\% & \textbf{50.20}\% \\  \hline
  		\end{tabular}
  		}
\end{table}

As seen in Table~\ref{tbl:result}, we found that AFT alone could only provide standard accuracy similar to that of AT.
This is because AFT does not explicitly impose constraints on the representation, and the DNN's representations of clean examples gradually diverge from the original representations of the standardly pretrained DNN during AFT.
However, AFT with RGKD or NR increased the standard accuracy.
These results demonstrate that these constraint techniques help the DNN to preserve the representation during AFT and effectively mitigate the accuracy-robustness tradeoff, as expected.
Furthermore, AFT with both RGKD and NR achieved the highest standard accuracy and robustness.
As described above, the three key components of ARREST, \ie, AFT, RGKD, and NR, work complementarily to obtain suitable representations of both clean and adversarial examples.
Therefore, the ablation study results indicate that the components' complementary roles actually serve to mitigate the accuracy-robustness tradeoff.
Finally, ARREST achieved a standard accuracy of 2.0 to 3.0 points higher than that of AT while also demonstrating higher adversarial robustness.
Overall, these results suggest that ARREST successfully mitigates the accuracy-robustness tradeoff in AT.

\section{Analysis of ARREST}
\label{sect:anal}

\begin{figure}[t]
  \begin{minipage}[b]{0.49\columnwidth}
    \centering
    \includegraphics[width=41mm]{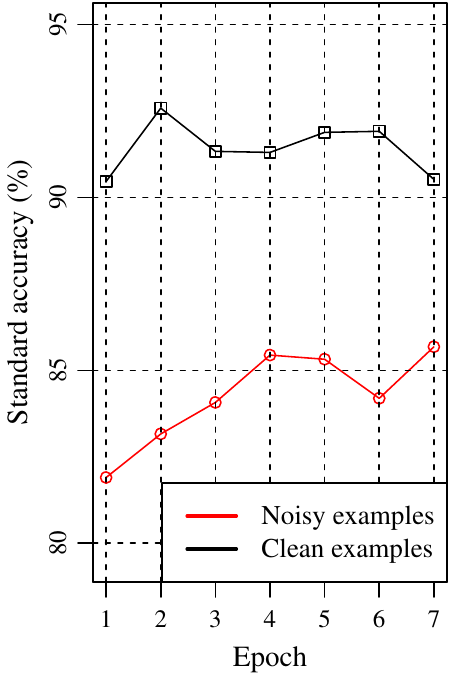}
  \end{minipage}~~
  \begin{minipage}[b]{0.49\columnwidth}
    \centering
    \includegraphics[width=41mm]{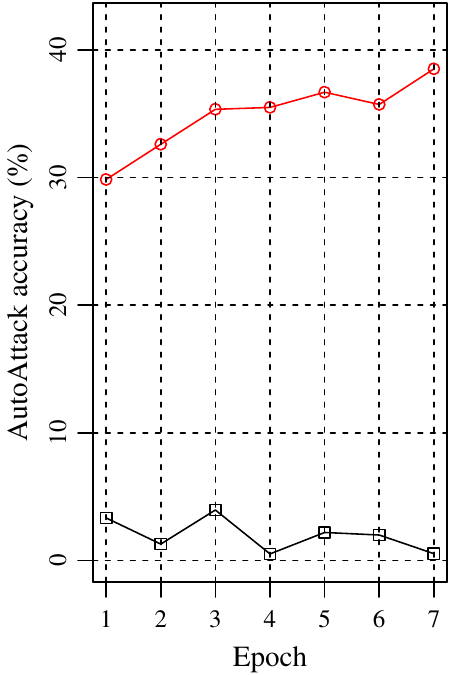}
  \end{minipage}
  \caption{Left: Relationship between the number of epochs and the standard accuracy in AFT with the replay technique.
    Right: Relationship between the number of epochs and the adversarial robustness with the same DNN used on the left.
    The black and red lines show results obtained by inputting clean and adversarial examples, respectively.
  }
    \label{fig:robust_nr}
\end{figure}

In this section, we analyze ARREST from four viewpoints: noisy examples in NR, the effect of ARREST on preserving the representation, comparison of other types of knowledge distillation with RGKD, and the effect of additional examples on ARREST.

\begin{figure}[t]
  \centering
   \includegraphics[width=90mm]{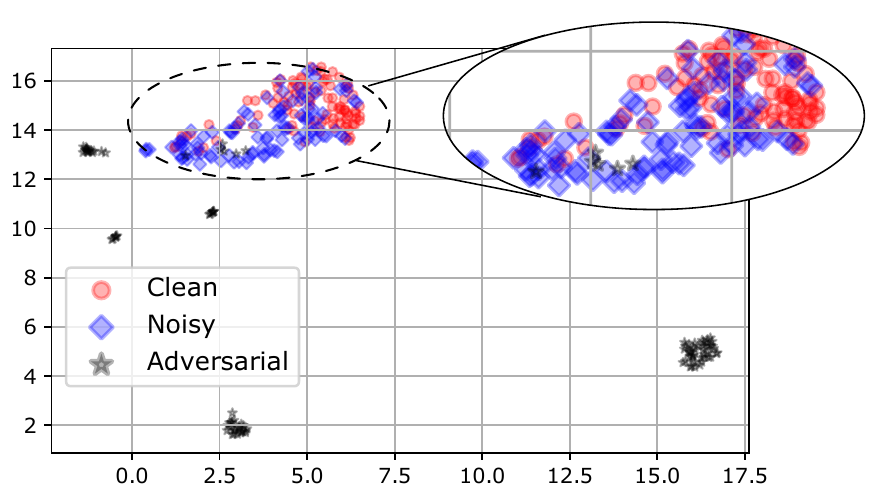}
  \caption{Visualization of representations of clean, noisy, and adversarial examples on a standardly pretrained DNN by using UMAP~\cite{McInnes18}.
    From CIFAR-10, 100 randomly selected test images labeled ``dog'' were used.
    The solid ellipse shows an enlargement of the dashed region.
  }
  \label{fig:tsne_nr}
\end{figure}

\noindent
\textbf{Effect of Inputting Noisy Examples on NR.}
First, we analyzed the effect of random noisy examples on the NR performance.
As mentioned in Subsection~\ref{subsect:nr}, we could consider inputting a clean example in Eq.~(\ref{formula:nr}); however, we found that this did not improve the adversarial robustness during AFT at all, as shown on the right in Fig.~\ref{fig:robust_nr}.
This is because partial input of clean examples during AFT makes optimization challenging with the distribution mismatch between clean and adversarial examples.
As a result, the DNN only obtains suitable representations of clean examples.
Actually, it maintains a high standard accuracy, as shown on the left in Fig.~\ref{fig:robust_nr}.
To avoid this issue, we use noisy examples in NR.
Figure~\ref{fig:tsne_nr} shows a visualization of the representations of clean, noisy, and adversarial examples with a standardly pretrained DNN.
Because the uniform noise is nonadversarial, the distribution underlying noisy examples is similar to that underlying clean examples but shifted slightly toward that of adversarial examples.
Hence, the use of noisy examples is expected to alleviate the distribution mismatch while enabling the DNN to benefit from the replay technique.
In fact, replay with noisy examples, \ie, our proposed NR, can improve the robustness in contrast with the use of clean examples, as shown by the red line on the right in Fig.~\ref{fig:robust_nr}.

\begin{table}[tb]
	\centering
		\caption{Comparison of the effectiveness of each ARREST component in preserving representation, with the cosine similarity as the metric.}
        \label{tbl:cossim}
  		\begin{tabular}{lccccccc}  \hline
	               & Cosine similarity  \\ \hline\hline
	   AT         &  0.255 \\ \hdashline
	   AFT        &  0.753\\ 
	   AFT + RGKD &  0.894\\
	   AFT + NR   &  0.764 \\  
          ARREST     &  \textbf{0.901}\\\hline
  		\end{tabular}
\end{table}

\begin{figure}[t]
  \begin{minipage}[b]{0.49\columnwidth}
    \centering
    \includegraphics[width=43.5mm]{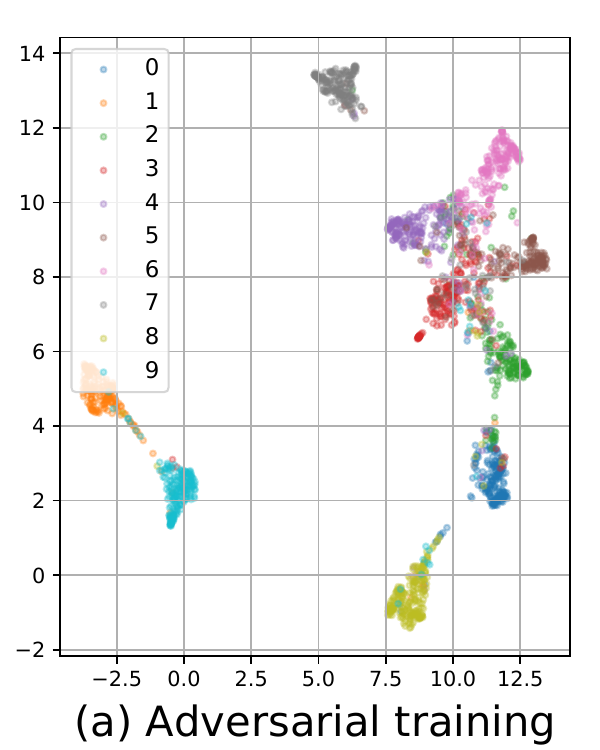}
  \end{minipage}
  \begin{minipage}[b]{0.49\columnwidth}
    \centering
    \includegraphics[width=43.5mm]{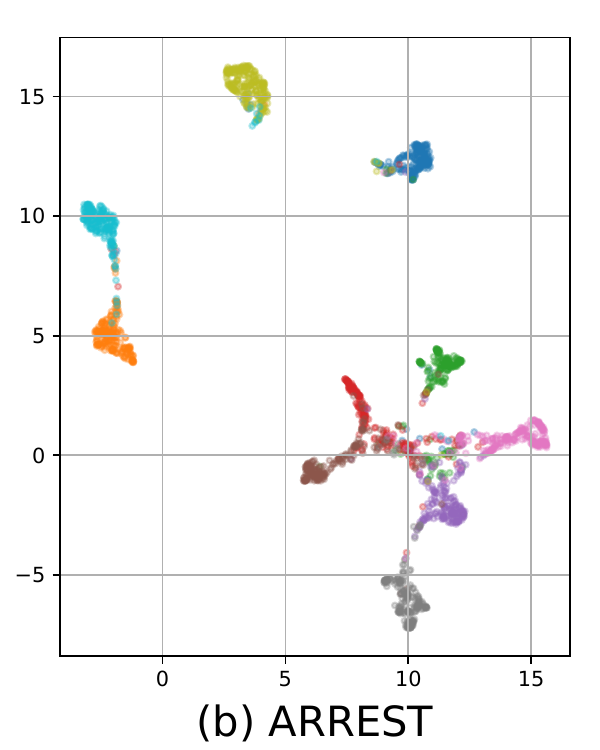}
  \end{minipage}
  \vspace{0.5mm}
  \caption{Visualization of clean example representations for (a)~a robust DNN with AT and (b)~ARREST by using UMAP~\cite{McInnes18}.
  For visualization, 2,000 randomly selected test images from CIFAR-10 were used.
  }
    \label{fig:tsne_comp}
\end{figure}

\noindent
\textbf{Effect of ARREST on Preserving Representation.}
Next, we analyzed the effectiveness of ARREST on preserving the representation from a standardly trained DNN.
ARREST's three key components were designed to obtain suitable representations of clean and adversarial examples by preserving the representation of clean examples from a standardly trained DNN.
To evaluate ARREST's proper operation, we compared the representations of robust and standardly pretrained DNNs.
Table~\ref{tbl:cossim} lists the cosine similarity between $h(\bm{x}; \theta_{\rm r})$ and $h(\bm{x}; \theta_{\rm s}^*)$ for evaluation on clean test examples from CIFAR-10.
As seen in the table, AT obtained a significantly different representation from that of the standardly trained DNN.
As explained above, this is due to the distribution mismatch issue~\cite{Xie20,Stutz19}.
AFT significantly increased the cosine similarity as compared with AT.
Moreover, AFT still had a gap between $h(\bm{x}; \theta_{\rm r})$ and $h(\bm{x}; \theta_{\rm s}^*)$ because of the remaining distribution mismatch, but each of RGKD and NR further increased the similarity.
Consequently, ARREST achieved the highest cosine similarity of 0.901.
Furthermore, as shown in Fig.~\ref{fig:tsne_comp}, we visualized the representations of clean examples with AT and ARREST.
We can see that ARREST~(b) obtained more discriminative representations of clean examples than AT did~(a).
These results indicate that preservation of the representation from the standardly trained DNN leads to higher standard accuracy, as expected.

\noindent
\textbf{Comparison with Other Types of Knowledge Distillation.}
As described in Subsection~\ref{subsetc:rgkd}, several AT methods~\cite{Cui21,Wang21} have applied knowledge distillation methodologies besides RGKD.
Here, we compared RGKD with them by replacing the loss in Eq.~(\ref{formula:2}) during AFT with two different methods.
First, ``logit''~\cite{Cui21} guides the DNN with the logit~(final output) of the pretrained DNN.
Second, ``attention map''~\cite{Wang21} guides the DNN with a spatial attention map computed by summing the latent representations along the channel dimension~\cite{Zagoruyko17}.
Appendix~\ref{sec:app_kd_other} provides these methods' detailed formulation and optimization.
Table~\ref{tbl:comp_const} lists the comparison results, which show that RGKD achieved the best performance in terms of both accuracy and robustness.
This may be because the comparison methods had a weak penalty effect by transferring the latent representation to a logit or attention map, in contrast with RGKD, which guides the DNN by using the representation as is.
For example, guiding with a logit does not directly affect the DNN representation, thus limiting the penalty effect.
This comparison indicates that RGKD is more suitable than the other methods for our aim of mitigating the accuracy-robustness tradeoff.

\begin{table}[tb]
	\centering
		\caption{Comparison of RGKD with other types of knowledge distillation.}
        \label{tbl:comp_const}
  		\begin{tabular}{lcccccc}  \hline
                              & Standard & AutoAttack \\ \hline\hline
Logit~\cite{Cui21}          &  87.45\%  & 49.62\% \\
Attention map~\cite{Wang21}  &  87.70\%  & 49.05\%\\ 
	   RGKD               &  \textbf{88.52}\%  & \textbf{50.20}\%\\ \hline
  		\end{tabular}
\end{table}

\noindent
\textbf{Effect of Additional Examples on ARREST.}
Finally, we investigated the effect of additional examples on ARREST, an approach that has often been used in previous works~\cite{Raghunathan20,Rade21,Gowal21} to mitigate the tradeoff.
Specifically, we used the additional examples from RST~\cite{Raghunathan20}, which were originally from~\cite{Carmon19}.
Appendix~\ref{sec:app_add_data} explains the training setup for this experiment.
The results are listed in Table~\ref{tbl:RST}.
ARREST with additional examples achieved higher standard accuracy and adversarial robustness than RST, which adversarially trains the DNN in the usual manner.
These results indicate that ARREST can benefit from additional examples, and the combination is promising for mitigating the accuracy-robustness tradeoff.

\section{Conclusion}

We have proposed AdversaRial finetuning with REpresentation conSTraint~(\textbf{ARREST}) to mitigate the accuracy-robustness tradeoff in adversarial training~(AT).
ARREST aims to obtain suitable representations of adversarial examples while preserving suitable representations of clean examples from standardly trained DNNs.
To this end, ARREST comprises three key components: (i)~\textit{adversarial finetuning}~(AFT), (ii)~\textit{representation-guided knowledge distillation}~(RGKD), and (iii)~\textit{noisy replay}~(NR).
It uses a two-step training process to obtain robust DNNs, entailing standard pretraining of DNNs on clean examples and finetuning of the pretrained DNNs on adversarial examples with RGKD and NR.
Further, we have proposed a new quantitative evaluation metric, accuracy robustness distance~(ARDist), which was inspired by the BD-Rate~\cite{Bjontegaard01,Strom20} metric used in video compression research.
Using ARDist, we demonstrated the quantitative effectiveness of ARREST in mitigating the tradeoff on CIFAR-10 and CIFAR-100 datasets.

\begin{table}[tb]
	\centering
		\caption{Results for ARREST with additional examples from RST~\cite{Raghunathan20}.}
        \label{tbl:RST}
  		\begin{tabular}{lccccccc}  \hline
	                & Standard & AutoAttack\\ \hline\hline
	   RST~\cite{Raghunathan20}  &  91.62\%   & 54.81\%\\ 
          ARREST w/ RST &  \textbf{93.17}\% & \textbf{55.73}\% \\\hline
  		\end{tabular}
\end{table}

While ARREST efficiently mitigates the accuracy-robustness tradeoff in AT, it could not perfectly eliminate the tradeoff; that is, it could not achieve the same standard accuracy as a standardly trained DNN.
A promising direction would be to combine ARREST with additional training examples, as was done to obtain the results in Table~\ref{tbl:RST}.
This might enable us to maximize the tradeoff mitigation or completely eliminate the tradeoff.
We plan to explore this direction in the future.

{\small
\bibliographystyle{ieee_fullname}
\bibliography{suzuki-cvpr}

\begin{thebibliography}{10}\itemsep=-1pt

\bibitem{Andriushchenko20}
M. Andriushchenko, F. Croce, N. Flammarion, and M. Hein.
\newblock Square attack: a query-efficient black-box adversarial attack via
  random search.
\newblock In {\em European Conference on Computer Vision~(ECCV)}, 2020.

\bibitem{Bengio09}
Y. Bengio, J. Louradour, R. Collobert, and J. Weston.
\newblock Curriculum learning.
\newblock In {\em International Conference on Machine Learning~(ICML)}, 2009.

\bibitem{Bjontegaard01}
G. Bj{\o}ntegaard.
\newblock Calculation of average {PSNR} differences between {RD-Curves}.
\newblock {\em ITU-T Video Coding Experts Group (VCEG)-M33}, 2001.

\bibitem{Boschini22}
M. Boschini, L. Bonicelli, A. Porrello, G. Bellitto, M. Pennisi, S. Palazzo, C.
  Spampinato, and S. Calderara.
\newblock Transfer without forgetting.
\newblock In {\em European Conference on Computer Vision~(ECCV)}, 2022.

\bibitem{Cai18}
Q. Cai, C. Liu, and D. Song.
\newblock Curriculum adversarial training.
\newblock In {\em International Joint Conference on Artificial
  Intelligence~(IJCAI)}, 2018.

\bibitem{Carlini17}
N. Carlini and D.~A. Wagner.
\newblock Towards evaluating the robustness of neural networks.
\newblock In {\em IEEE Symposium on Security and Privacy~(SP)}, 2017.

\bibitem{Carmon19}
Y. Carmon, A. Raghunathan, L. Schmidt, J.~C. Duchi, and P.~S. Liang.
\newblock Unlabeled data improves adversarial robustness.
\newblock In {\em Advances in Neural Information Processing Systems~(NeurIPS)},
  2019.

\bibitem{Chen22}
J. Chen, Y. Cheng, Z. Gan, Q. Gu, and J. Liu.
\newblock Efficient robust training via backward smoothing.
\newblock In {\em AAAI Conference on Artificial Intelligence~(AAAI)}, 2022.

\bibitem{Croce21}
F. Croce, M. Andriushchenko, V. Sehwag, E. Debenedetti, N. Flammarion, M.
  Chiang, P. Mittal, and M. Hein.
\newblock Robustbench: a standardized adversarial robustness benchmark.
\newblock In {\em Advances in Neural Information Processing Systems~(NeurIPS)},
  2021.

\bibitem{Croce20_FAB}
F. Croce and M. Hein.
\newblock Minimally distorted adversarial examples with a fast adaptive
  boundary attack.
\newblock In {\em International Conference on Machine Learning~(ICML)}, 2020.

\bibitem{Croce20}
F. Croce and M. Hein.
\newblock Reliable evaluation of adversarial robustness with an ensemble of
  diverse parameter-free attacks.
\newblock In {\em International Conference on Machine Learning~(ICML)}, 2020.

\bibitem{Cui21}
J. Cui, S. Liu, L. Wang, and J. Jia.
\newblock Learnable boundary guided adversarial training.
\newblock In {\em International Conference on Computer Vision~(ICCV)}, 2021.

\bibitem{Farquhar18}
S. Farquhar and Y. Gal.
\newblock Towards robust evaluations of continual learning.
\newblock {\em arXiv preprint arXiv:1805.09733}, 2018.

\bibitem{Goodfellow14}
I. Goodfellow, J. Pouget-Abadie, M. Mirza, B. Xu, D. Warde-Farley, S. Ozair, A.
  Courville, and Y. Bengio.
\newblock Generative adversarial nets.
\newblock In {\em Advances in Neural Information Processing Systems~(NIPS)},
  2014.

\bibitem{Goodfellow15}
I.~J. Goodfellow, J. Shlens, and C. Szegedy.
\newblock Explaining and harnessing adversarial examples.
\newblock In {\em International Conference on Learning Representations~(ICLR)},
  2015.

\bibitem{Gowal21}
S. Gowal, S.-A. Rebuffi, O. Wiles, F. Stimberg, D.~A. Calian, and T. Mann.
\newblock Improving robustness using generated data.
\newblock In {\em Advances in Neural Information Processing Systems~(NeurIPS)},
  2021.

\bibitem{Guo18}
C. Guo, M. Rana, M. Cisse, and L. van~der Maaten.
\newblock Countering adversarial images using input transformations.
\newblock In {\em International Conference on Learning Representations~(ICLR)},
  2018.

\bibitem{He16}
K. He, X. Zhang, S. Ren, and J. Sun.
\newblock Deep residual learning for image recognition.
\newblock In {\em IEEE Conference on Computer Vision and Pattern
  Recognition~(CVPR)}, 2016.

\bibitem{Hinton15_distillation}
G.~E. Hinton, O. Vinyals, and J. Dean.
\newblock Distilling the knowledge in a neural network.
\newblock In {\em NIPS Deep Learning and Representation Learning Workshop},
  2015.

\bibitem{Ilyas19}
A. Ilyas, S. Santurkar, D. Tsipras, L. Engstrom, B. Tran, and A. M\k{a}dry.
\newblock Adversarial examples are not bugs, they are features.
\newblock In {\em Advances in Neural Information Processing Systems~(NeurIPS)},
  2019.

\bibitem{Ioffe15}
S. Ioffe and C. Szegedy.
\newblock Batch normalization: Accelerating deep network training by reducing
  internal covariate shift.
\newblock In {\em International Conference on Machine Learning~(ICML)}, 2015.

\bibitem{Jeddi20}
A. Jeddi, M.~J. Shafiee, and A. Wong.
\newblock A simple fine-tuning is all you need: Towards robust deep learning
  via adversarial fine-tuning.
\newblock {\em arXiv preprint, arXiv:2012.13628}, 2020.

\bibitem{Jia22}
X. Jia, Y. Zhang, B. Wu, J.~Wang K.~Ma, and X. Cao.
\newblock {LAS-AT}: Adversarial training with learnable attack strategy.
\newblock In {\em IEEE Conference on Computer Vision and Pattern
  Recognition~(CVPR)}, 2022.

\bibitem{Jin22}
G. Jin, X. Yi, W. Huang, S. Schewe, and X. Huang.
\newblock Enhancing adversarial training with second-order statistics of
  weights.
\newblock In {\em IEEE Conference on Computer Vision and Pattern
  Recognition~(CVPR)}, 2022.

\bibitem{Kannan18}
H. Kannan, A. Kurakin, and I. Goodfellow.
\newblock Adversarial logit pairing.
\newblock {\em arXiv preprint, arXiv:1803.06373}, 2018.

\bibitem{Kim20}
J. Kim and X. Wang.
\newblock Sensible adversarial learning.
\newblock In {\em OpenReview}, 2019.

\bibitem{Krizhevsky09}
A. Krizhevsky.
\newblock Learning multiple layers of features from tiny images.
\newblock Technical report, University of Toronto, 2009.

\bibitem{Krizhevsky12}
A. Krizhevsky, I. Sutskever, and G.~E. Hinton.
\newblock Imagenet classification with deep convolutional neural networks.
\newblock In {\em Advances in Neural Information Processing Systems~(NIPS)},
  2012.

\bibitem{Kumari19}
N. Kumari, M. Singh, A. Sinha, H. Machiraju, B. Krishnamurthy, and V.~N
  Balasubramanian.
\newblock Harnessing the vulnerability of latent layers in adversarially
  trained models.
\newblock In {\em International Joint Conference on Artificial
  Intelligence~(IJCAI)}, 2019.

\bibitem{Lee18}
K. Lee, K. Lee, H. Lee, and J. Shin.
\newblock A simple unified framework for detecting out-of-distribution samples
  and adversarial attacks.
\newblock In {\em Advances in Neural Information Processing Systems~(NeurIPS)},
  2018.

\bibitem{Li23}
Q. Li, Y. Guo, W. Zuo, and H. Chen.
\newblock Squeeze training for adversarial robustness.
\newblock In {\em International Conference on Learning Representations~(ICLR)},
  2023.

\bibitem{Long15}
J. Long, E. Shelhamer, and T. Darrell.
\newblock Fully convolutional networks for semantic segmentation.
\newblock In {\em IEEE Conference on Computer Vision and Pattern
  Recognition~(CVPR)}, 2015.

\bibitem{Ma18}
X. Ma, B. Li, Y. Wang, S.~M. Erfani, S. Wijewickrema, G. Schoenebeck, M.~E.
  Houle, D. Song, and J. Bailey.
\newblock Characterizing adversarial subspaces using local intrinsic
  dimensionality.
\newblock In {\em International Conference on Learning Representations~(ICLR)},
  2018.

\bibitem{Mao19}
C. Mao, Z. Zhong, J. Yang, C. Vondrick, and B. Ray.
\newblock Metric learning for adversarial robustness.
\newblock In {\em Advances in Neural Information Processing Systems~(NeurIPS)},
  2019.

\bibitem{Mccloskey89}
M. Mccloskey and N.~J. Cohen.
\newblock Catastrophic interference in connectionist networks: {T}he sequential
  learning problem.
\newblock {\em The Psychology of Learning and Motivation}, 24:104--169, 1989.

\bibitem{McInnes18}
L. McInnes, J. Healy, N. Saul, and L. Grossberger.
\newblock {UMAP}: Uniform manifold approximation and projection.
\newblock {\em The Journal of Open Source Software}, 3(29):861, 2018.

\bibitem{Madry18}
A. M\k{a}dry, A. Makelov, L. Schmidt, D. Tsipras, and A. Vladu.
\newblock Towards deep learning models resistant to adversarial attacks.
\newblock In {\em International Conference on Learning Representations~(ICLR)},
  2018.

\bibitem{Moosavi-Dezfooli16}
S. Moosavi{-}Dezfooli, A. Fawzi, and P. Frossard.
\newblock Deepfool: {A} simple and accurate method to fool deep neural
  networks.
\newblock In {\em {IEEE} Conference on Computer Vision and Pattern
  Recognition~(CVPR)}, 2016.

\bibitem{Moosavi-Dezfooli19}
S. Moosavi-Dezfooli, A. Fawzi, J. Uesato, and P. Frossard.
\newblock Robustness via curvature regularization, and vice versa.
\newblock In {\em IEEE Conference on Computer Vision and Pattern
  Recognition~(CVPR)}, 2019.

\bibitem{Pang21}
T. Pang, X. Yang, Y. Dong, H. Su, and J. Zhu.
\newblock Bag of tricks for adversarial training.
\newblock In {\em International Conference on Learning Representations~(ICLR)},
  2021.

\bibitem{Papernot17}
N. Papernot, P. McDaniel, I. Goodfellow, S. Jha, Z.~B. Celik, and A. Swami.
\newblock Practical black-box attacks against machine learning.
\newblock In {\em Asia Conference on Computer and Communications
  Security~(CCS)}, 2017.

\bibitem{Rade21}
R. Rade and S. Moosavi-Dezfooli.
\newblock Helper-based adversarial training: Reducing excessive margin to
  achieve a better accuracy vs. robustness trade-off.
\newblock In {\em International Conference on Machine Learning~(ICML) Workshop
  on Adversarial Machine Learning}, 2021.

\bibitem{Raghunathan20}
A. Raghunathan, S.~M. Xie, F. Yang, J.~C. Duchi, and P. Liang.
\newblock Understanding and mitigating the tradeoff between robustness and
  accuracy.
\newblock In {\em International Conference on Machine Learning~(ICML)}, 2020.

\bibitem{Ramasesh21}
V.~V. Ramasesh, E. Dyer, and M. Raghu.
\newblock Anatomy of catastrophic forgetting: Hidden representations and task
  semantics.
\newblock In {\em Advances in Neural Information Processing Systems~(NeurIPS)},
  2021.

\bibitem{Redmon17}
J. Redmon and A. Farhadi.
\newblock {YOLO9000:} better, faster, stronger.
\newblock In {\em IEEE Conference on Computer Vision and Pattern
  Recognition~(CVPR)}, 2017.

\bibitem{Rice20}
L. Rice, E. Wong, and Z. Kolter.
\newblock Overfitting in adversarially robust deep learning.
\newblock In {\em International Conference on Machine Learning~(ICML)}, 2020.

\bibitem{Rolnick19}
D. Rolnick, A. Ahuja, J. Schwarz, T. Lillicrap, and G. Wayne.
\newblock Experience replay for continual learning.
\newblock In {\em Advances in Neural Information Processing Systems~(NeurIPS)},
  2019.

\bibitem{Romero15}
A. Romero, N. Ballas, S.~E. Kahou, A. Chassang, C. Gatta, and Y. Bengio.
\newblock Fitnets: Hints for thin deep nets.
\newblock In {\em International Conference on Learning Representations~(ICLR)},
  2015.

\bibitem{Samangouei18}
P. Samangouei, M. Kabkab, and R. Chellappa.
\newblock Defense-{GAN}: Protecting classifiers against adversarial attacks
  using generative models.
\newblock In {\em International Conference on Learning Representations~(ICLR)},
  2018.

\bibitem{Schmidt18}
L. Schmidt, S. Santurkar, D. Tsipras, K. Talwar, and A. M\k{a}dry.
\newblock Adversarially robust generalization requires more data.
\newblock In {\em Advances in Neural Information Processing Systems~(NeurIPS)},
  2018.

\bibitem{Simonyan14c_ICLR}
K. Simonyan and A. Zisserman.
\newblock Very deep convolutional networks for large-scale image recognition.
\newblock In {\em International Conference on Learning Representations~(ICLR)},
  2015.

\bibitem{Sitawarin21}
C. Sitawarin, S. Chakraborty, and D. Wagner.
\newblock Sat: Improving adversarial training via curriculum-based loss
  smoothing.
\newblock In {\em ACM Workshop on Artificial Intelligence and
  Security~(AISec)}, 2021.

\bibitem{Strom20}
J. Str\"om, K. Andersson, R. Sj\"oberg, F. Bossen, G. Sullivan, and J.-R. Ohm.
\newblock Summary information on bd-rate experiment evaluation practices.
\newblock {\em Joint Video Experts Team (JVET)-Q2016}, 2020.

\bibitem{Stutz19}
D. Stutz, M. Hein, and B. Schiele.
\newblock Disentangling adversarial robustness and generalization.
\newblock In {\em IEEE Conference on Computer Vision and Pattern
  Recognition~(CVPR)}, 2019.

\bibitem{Szegedy14}
C. Szegedy, W. Liu, Y. Jia, P. Sermanet, S. Reed, D. Anguelov, D. Erhan, V.
  Vanhoucke, and A. Rabinovich.
\newblock Going deeper with convolutions.
\newblock In {\em IEEE Conference on Computer Vision and Pattern
  Recognition~(CVPR)}, 2015.

\bibitem{Szegedy14b}
C. Szegedy, W. Zaremba, I. Sutskever, J. Bruna, D Erhan, I.~J. Goodfellow, and
  R. Fergus.
\newblock Intriguing properties of neural networks.
\newblock In {\em International Conference on Learning Representations~(ICLR)},
  2014.

\bibitem{Tsipras18}
D. Tsipras, S. Santurkar, L. Engstrom, A. Turner, and A. M\k{a}dry.
\newblock Robustness may be at odds with accuracy.
\newblock In {\em International Conference on Learning Representations~(ICLR)},
  2019.

\bibitem{Wang21}
H. Wang, Y. Deng, S. Yoo, H. Ling, and Y. Lin.
\newblock {AGKD-BML:} defense against adversarial attack by attention guided
  knowledge distillation and bi-directional metric learning.
\newblock In {\em International Conference on Computer Vision~(ICCV)}, 2021.

\bibitem{Wang19_BAT}
J. Wang and H. Zhang.
\newblock Bilateral adversarial training: Towards fast training of more robust
  models against adversarial attacks.
\newblock In {\em International Conference on Computer Vision~(ICCV)}, 2019.

\bibitem{Wang17}
J. Wang, F. Zhou, S. Wen, X. Liu, and Y. Lin.
\newblock Deep metric learning with angular loss.
\newblock In {\em International Conference on Computer Vision~(ICCV)}, 2017.

\bibitem{Wang19_DAT}
Y. Wang, X. Ma, J. Bailey, J. Yi, B. Zhou, and Q. Gu.
\newblock On the convergence and robustness of adversarial training.
\newblock In {\em International Conference on Machine Learning~(ICML)}, 2019.

\bibitem{Wang20}
Y. Wang, D. Zou, J. Yi, J. Bailey, X. Ma, and Q. Gu.
\newblock Improving adversarial robustness requires revisiting misclassified
  examples.
\newblock In {\em International Conference on Learning Representations~(ICLR)},
  2020.

\bibitem{Wu20}
D. Wu, S. Xia, and Y. Wang.
\newblock Adversarial weight perturbation helps robust generalization.
\newblock In {\em Advances in Neural Information Processing Systems~(NeurIPS)},
  2020.

\bibitem{Xie20}
C. Xie, M. Tan, B. Gong, J. Wang, A.~L. Yuille, and Q.~V. Le.
\newblock Adversarial examples improve image recognition.
\newblock In {\em IEEE Conference on Computer Vision and Pattern
  Recognition~(CVPR)}, 2020.

\bibitem{Xu17}
W. Xu, D. Evans, and Y. Qi.
\newblock Feature squeezing: Detecting adversarial examples in deep neural
  networks.
\newblock {\em arXiv preprint arXiv:1704.01155}, 2017.

\bibitem{Zagoruyko16}
S. Zagoruyko and N. Komodakis.
\newblock Wide residual networks.
\newblock In {\em British Machine Vision Conference~(BMVC)}, 2016.

\bibitem{Zagoruyko17}
S. Zagoruyko and N. Komodakis.
\newblock Paying more attention to attention: Improving the performance of
  convolutional neural networks via attention transfer.
\newblock In {\em International Conference on Learning Representations~(ICLR)},
  2017.

\bibitem{Zhang19_FS}
H. Zhang and J. Wang.
\newblock Defense against adversarial attacks using feature scattering-based
  adversarial training.
\newblock In {\em Advances in Neural Information Processing Systems~(NeurIPS)},
  2019.

\bibitem{Zhang20_AIT}
H. Zhang and W. Xu.
\newblock Adversarial interpolation training: A simple approach for improving
  model robustness.
\newblock In {\em OpenReview}, 2020.

\bibitem{Zhang19_TRADES}
H. Zhang, Y. Yu, J. Jiao, E.~P. Xing, L.~E. Ghaoui, and M.~I. Jordan.
\newblock Theoretically principled trade-off between robustness and accuracy.
\newblock In {\em International Conference on Machine Learning~(ICML)}, 2019.

\bibitem{Zhang20_FAT}
J. Zhang, X. Xu, B. Han, G. Niu, L. Cui, M. Sugiyama, and M. Kankanhalli.
\newblock Attacks which do not kill training make adversarial learning
  stronger.
\newblock In {\em International Conference on Machine Learning~(ICML)}, 2020.

\bibitem{Zhu22}
J. Zhu, J. Yao, B. Han, J. Zhang, T. Liu, G. Niu, J. Zhou, J. Xu, and H. Yang.
\newblock Reliable adversarial distillation with unreliable teachers.
\newblock In {\em International Conference on Learning Representations~(ICLR)},
  2022.

\end{thebibliography}
}

\clearpage

\onecolumn

\section*{{\LARGE Supplementary Material}}

\appendix

\section{Detail of Existing Methods of Comparison Experiments}
\label{sec:app_exist}
In order to evaluate the performance of ARREST, we selected many existing methods for comparison.
Table~\ref{tbl:app_model_c10-100} lists the DNNs trained on existing methods, with 22 and 13 DNN models for CIFAR-10 and CIFAR-100, respectively.
The standard and AutoAttack accuracies, Sum, ARDist, and sources of their accuracies are depicted.
Almost all models used WideResNet-34-10 as an architecture, though several ones used WideResNet-28-10.
We found that AWP~\cite{Wu20}, LAS-AT~\cite{Jia22}, S$^2$O~\cite{Jin22}, and LBGAT~\cite{Cui21} had the four best scores in terms of Sum and ARDist on both CIFAR-10 and CIFAR-100.

\begin{table*}[htb]
\centering
\caption{Detailed information on existing methods.
The source of each accuracy is also described in the rightmost column.
The methods are sorted using the score of ARDist in ascending order.
* indicates a result obtained with WideResNet-28-10; the other results were obtained with WideResNet-34-10.
}
\label{tbl:app_model_c10-100}
\begin{tabular}{l|cc|cc|l}
                                & Standard & AutoAttack & Sum    & ARDist & Source of performance \\ \hline
\multicolumn{6}{c}{\textbf{CIFAR-10} - $\varepsilon = 8/255$} \\ \hline 
AT~\cite{Madry18}               & 87.14\%  & 44.04\%    & 131.18 & -1.500 & RobustBench\\ \hdashline
DAT~\cite{Wang19_DAT}           & 86.20\%  & 45.38\%    & 132.10 & -1.991 & Directly copied from~\cite{Sitawarin21} \\
TLA~\cite{Mao19}                & 86.21\%  & 47.41\%    & 133.62 & -1.282 & RobustBench\\
CAT~\cite{Cai18}                & 89.61\%  & 34.78\%    & 124.39 & -1.259 & Directly copied from~\cite{Sitawarin21} \\
MART~\cite{Wang20}              & 83.62\%  & 50.98\%    & 134.60 & -0.922 & Provided model\\
BAT~\cite{Wang19_BAT}*          & 91.20\%  & 29.35\%    & 120.55 & -0.690 & RobustBench\\
FS~\cite{Zhang19_FS}*           & 90.00\%  & 36.64\%    & 126.64 & -0.502 & RobustBench\\
AIT~\cite{Zhang20_AIT}*         & 90.25\%  & 36.45\%    & 126.70 & -0.297 & RobustBench\\
BS~\cite{Chen22}                & 85.32\%  & 51.12\%    & 136.44 & -0.230 & RobustBench\\
AGKD-BML~\cite{Wang21}*         & 86.25\%  & 50.59\%    & 136.84 & 0.184  & Directly copied from~\cite{Wang21}\\
FAT~\cite{Zhang20_FAT}          & 89.34\%  & 43.05\%    & 132.39 & 0.343  & Provided model\\
SAL~\cite{Kim20}                & 91.51\%  & 34.22\%    & 125.73 & 0.496  & RobustBench\\
IAD~\cite{Zhu22}                & 85.09\%  & 52.29\%    & 137.38 & 0.535  & Directly copied from~\cite{Zhu22}\\
SAT~\cite{Sitawarin21}          & 86.84\%  & 50.75\%    & 137.59 & 0.766  & Directly copied from~\cite{Sitawarin21}\\
LAT~\cite{Kumari19}             & 87.80\%  & 49.12\%    & 136.92 & 0.853  & RobustBench\\
TRADES~\cite{Zhang19_TRADES}    & 84.92\%  & 53.08\%    & 138.00 & 1.180  & RobustBench\\
ST~\cite{Li23}                  & 84.92\%  & 53.54\%    & 138.46 & 1.616  & Directly copied from~\cite{Li23}\\
Bag of Tricks~\cite{Pang21}     & 84.24\%  & 53.88\%    & 138.12 & 1.829  & Directly copied from~\cite{Pang21}\\
LAS-AT~\cite{Jia22}             & 86.23\%  & 53.58\%    & 139.81 & 2.236  & Directly copied from~\cite{Jia22}\\ 
AWP~\cite{Wu20}                 & 85.57\%  & 54.04\%    & 139.61 & 2.314  & Directly copied from~\cite{Wu20}\\
S$^2$O~\cite{Jin22} & 85.67\%  & 54.10\%    & 139.77 & 2.410  & Directly copied from~\cite{Jin22}\\
LBGAT~\cite{Cui21}              & 88.22\%  & 52.18\%    & 140.40 & 2.706  & Provided model\\
\hline
\multicolumn{6}{c}{\textbf{CIFAR-100} - $\varepsilon = 8/255$} \\ \hline 
AT~\cite{Madry18}               & 59.59\%  & 22.86\%    & 82.45  & -3.268 & Our reimplementation\\ \hdashline
AIT~\cite{Zhang20_AIT}*         & 69.51\%  &  2.80\%    & 72.31  & -7.403 & Provided model\\
CAT~\cite{Cai18}                & 62.84\%  & 16.82\%    & 79.66  & -5.487 & Directly copied from~\cite{Sitawarin21} \\
FS~\cite{Zhang19_FS}*           & 73.94\%  &  0.04\%    & 73.98  & -4.728 & Provided model\\
FAT~\cite{Zhang20_FAT}          & 65.51\%  & 21.17\%    & 86.68  & -0.618 & Provided model\\
TRADES~\cite{Zhang19_TRADES}    & 56.50\%  & 26.87\%    & 83.37  & -0.449 & Directly copied from~\cite{Cui21}\\
SAT~\cite{Sitawarin21}          & 62.95\%  & 24.56\%    & 87.51  & 0.074  & Directly copied from~\cite{Sitawarin21}\\
BS~\cite{Chen22}                & 62.15\%  & 26.94\%    & 89.09  & 1.549  & RobustBench\\
IAD~\cite{Zhu22}                & 60.72\%  & 27.89\%    & 88.61  & 1.687  & Directly copied from~\cite{Zhu22}\\
AWP~\cite{Wu20}                 & 60.38\%  & 28.86\%    & 89.24  & 2.424  & Directly copied from~\cite{Wu20}\\
S$^2$O~\cite{Jin22} & 63.40\%  & 27.60\%    & 91.00  & 2.786  & Directly copied from~\cite{Jin22}\\
LAS-AT~\cite{Jia22}             & 61.80\%  & 29.03\%    & 90.83  & 3.189  & Directly copied from~\cite{Jia22}\\
LBGAT~\cite{Cui21}              & 70.25\%  & 26.73\%    & 96.98  & 6.639  & Provided model\\
\end{tabular}
\end{table*}

\section{Detail of ARDist}
\label{sec:app_res-func}
As described in Subsection~\ref{subsect:metrics}, ARDist approximates a curve representing the accuracy-robustness tradeoff.
The red dashed lines in Figs.~\ref{fig:intro_res} and \ref{fig:res_100} show the approximated curves.
ARDist evaluates the mitigation by calculating the distance between the approximated curve and a point given by the method being evaluated.
We can calculate the distance by finding the approximated curve's normal through the evaluated point.
This can be done with numerical calculation.
The source codes for the calculation of ARDist are as follows.
One can easily use these codes by only setting the standard and AutoAttack accuracies of a method that is required evaluation.

\begin{mdframed}
\begin{minted}[fontsize=\small]{python}
## ARDist for CIFAR-10 ##
from scipy import optimize, exp
import numpy as np

p=90.24 # please set standard accuracy
q=50.20 # please set AutoAttack accuracy

def f(x):
    return 9.877e-05*x**3 - 0.3922*x**2 + 63.82*x - 2600

def df(x):
    return 9.877e-05*3*x**2 - 0.3922*2*x + 63.82

def h(x):
    return q-f(x) + (p-x)/df(x)

x_sol = float(optimize.fsolve(h,90))
sign = np.sign(q-f(p))
print("x = " + str(x_sol))
print("y = " + str(f(x_sol)))
print("Sign: " + str(sign))
print(sign * np.sqrt((p-x_sol)**2+(q-f(x_sol))**2))
\end{minted}
\end{mdframed}
\begin{mdframed}
\begin{minted}[fontsize=\small]{python}
## ARDist for CIFAR-100 ##
from scipy import optimize, exp
import numpy as np

p=73.05 # please set the standard accuracy
q=24.32 # please set the AutoAttack accuracy

def f(x):
    return 0.0005615*x**3 - 0.1582*x**2 + 12.44*x - 271.8

def df(x):
    return 0.0005615*3*x**2 - 0.1582*2*x + 12.44

def h(x):
    return q-f(x) + (p-x)/df(x)

x_sol = float(optimize.fsolve(h,70))
sign = np.sign(q-f(p))
print("x = " + str(x_sol))
print("y = " + str(f(x_sol)))
print("Sign: " + str(sign))
print(sign * np.sqrt((p-x_sol)**2+(q-f(x_sol))**2))
\end{minted}
\end{mdframed}

\section{Results with Other Distance Functions}
\label{sec:app_res_function}

In the experiment, we used the angular distance as $d(\cdot)$ in RGKD.
However, RGKD also performs well with other distance functions.
Table~\ref{tbl:app_comp_dist} lists the results of RGKD depending on the distance functions.
As shown, using the mean squared error~(MSE) and mean absolute error~(MAE) achieves almost the same performance of the angular distance.
Since using the angular distance achieves slightly better than the other two functions, we chose it in the experiments.

\begin{table}[htb]
	\centering
		\caption{Comparison of three distance functions in RGKD.}
        \label{tbl:app_comp_dist}
  		\begin{tabular}{lcccccc}  \hline
                        & Standard  & AutoAttack \\ \hline\hline
            MSE         &  88.34\%  & 50.06\% \\
            MAE         &  88.31\%  & 50.13\%\\ 
Angular distance        &  \textbf{88.52}\%  & \textbf{50.20}\%\\ \hline
  		\end{tabular}
\end{table}

\section{Comparison with TRADES and S$^2$O using Varied Hyperparameters}
\label{sec:app_trades_hypara}

\begin{figure}[htb]
\centering
 \begin{minipage}{0.375\hsize}
  \centering
   \includegraphics[width=50mm]{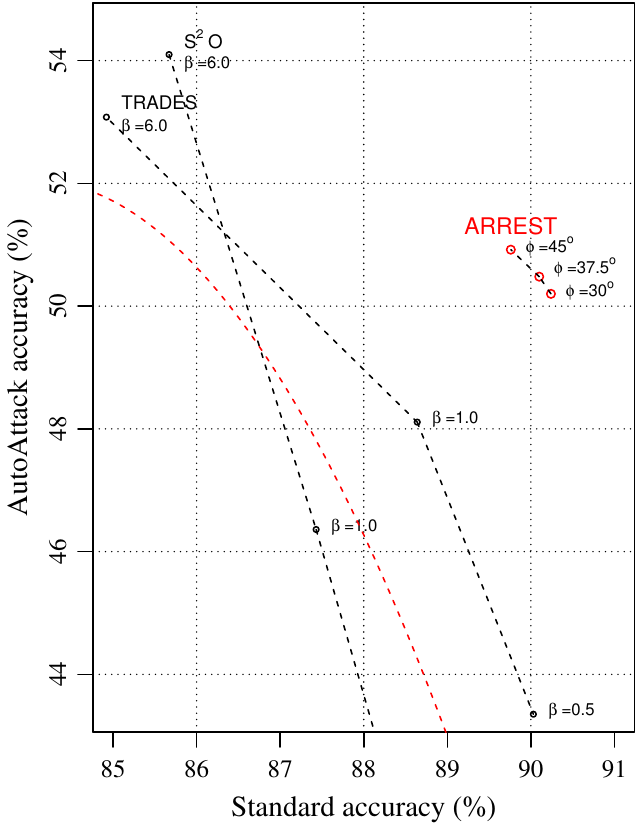}
  \caption{Comparison results between ARREST, TRADE, and S$^2$O for CIFAR-10.}
  \label{fig:app_trades_10}
 \end{minipage}~~~~~~~~~
 \begin{minipage}{0.45\hsize}
  \centering
   \includegraphics[width=72.5mm]{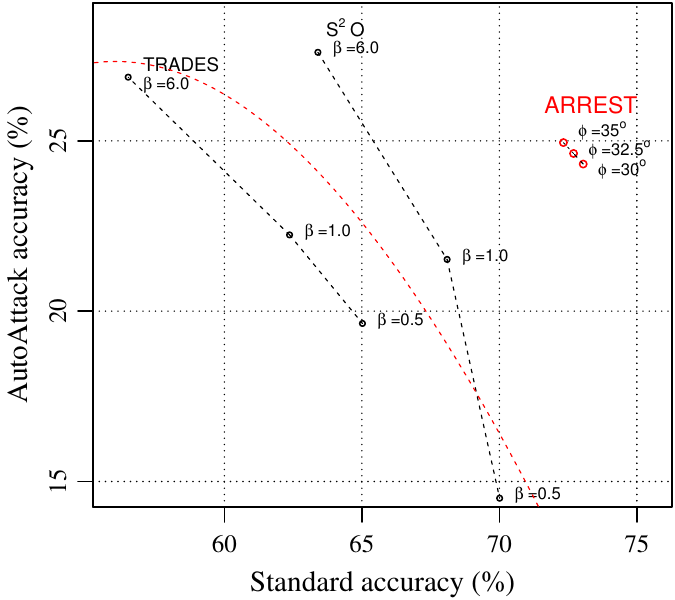}
  \caption{Comparison results between ARREST, TRADE, and S$^2$O for CIFAR-100.}
  \label{fig:app_trades_100}
 \end{minipage}
\end{figure}

As discussed in Section~\ref{sect:relate}, TRADES~\cite{Zhang19_TRADES} can achieve various accuracy-robustness tradeoffs by adjusting the hyperparameter $\beta$.
Although we used the results of TRADES with $\beta=6.0$ for comparison in Figs.~\ref{fig:intro_res} and \ref{fig:res_100}, it is also important to compare ARREST with TRADES when other hyperparameters are used.
Here, we compared ARREST with not only TRADES but also a recent state-of-the-art method called S$^2$O~\cite{Jin22}, which is integrated with TRADES.
We set three values $\{0.5, 1.0, 6.0\}$ as the hyperparameter $\beta$.
Figures~\ref{fig:app_trades_10} and \ref{fig:app_trades_100} show the results for CIFAR-10 and CIFAR-100, respectively.
In these figures, the results of ARREST clearly appear on the right side relative to those of TRADES and S$^2$O, thus indicating that ARREST can achieve higher standard accuracy while maintaining the same robustness as these methods.
Finally, in both figures, we also show the approximated curve of the tradeoff for ARDist, the same as in Figs.~\ref{fig:intro_res} and \ref{fig:res_100}.
As we can see, these curves are in good agreement with the results of ARREST and S$^2$O.

\section{Results of Other Attacks}
\label{sec:app_other_attack}

In Section~\ref{sec:exp}, we mainly evaluated the robustness using AutoAttack~\cite{Croce20} since it is common and reliable.
In this appendix, we evaluated the robustness of ARREST using other attacks.
Table~\ref{tbl:app_result} lists the results obtained with four variations of our method~(only AFT, AFT with RGKD, AFT with NR, and all of them, \ie, ARREST) and two baselines~(standard training and AT).
We evaluated them using four types of adversarial attacks: FGSM~\cite{Goodfellow15}, PGD~(7 and 30)~\cite{Madry18}, and CW~(30)~\cite{Carlini17}.
The number in parentheses indicates the number of iterative steps for each attack.
As seen in Table~\ref{tbl:app_result}, the robustness trend of the variations of our method is consistent with those in Table~\ref{tbl:result}.
Namely, ARREST also exhibits high adversarial robustness against various attacks, not only AutoAttack.

\begin{table*}[htb]
	\centering
		\caption{Results obtained with four variations of our method~(only AFT, AFT with RGKD, AFT with NR, ARREST) and two baselines~(standard training and AT).
        Evaluation of robustness is done with FGSM~\cite{Goodfellow15}, PGD~(7 and 30)~\cite{Madry18}, and CW~(30)~\cite{Carlini17}.
        Number in parentheses indicates number of iterative steps for each attack.
        }
        \label{tbl:app_result}
		\scalebox{0.9}{
  		\begin{tabular}{l|ccccc|cccccc}  \hline
              & \multicolumn{5}{c}{ResNet-18} & \multicolumn{5}{|c}{WideResNet-34-10} \\
	         & Standard    & FGSM     & PGD~(7)  & PGD~(20) & CW~(30) & Standard & FGSM     & PGD~(7)  & PGD~(20) & CW~(30) \\ \hline\hline
	   ST    & \textbf{94.5}\% & 28.3\% & 0\% & 0\% & 0\% & \textbf{95.4}\% & 38.6\% & 0\%     & 0\%     & 0\%\\
	   AT~\cite{Madry18}   & 84.7\% & 56.0\% & 50.6\% & 46.7\% & 46.9\% & 87.1\%  & 56.1\%  & 50.0\%  & 45.8\%  & 46.8\%\\ \hdashline
	   AFT & 84.1\% & 56.2\% & 52.4\% & 49.2\% & 48.0\% & 87.5\% & 60.0\% & 55.0\% & 51.3\% & 51.3\%\\
	   AFT + RGKD       & 85.1\% & 57.4\% & 53.0\% & \textbf{49.6}\% & 49.0\% & 88.5\% & 61.5\% & 56.0\% & 52.3\% & 52.9\%\\ 
	   AFT + NR        & 85.5\% & 56.5\% & 52.4\% & 49.0\% & 48.1\% & 88.9\% & 60.1\% & 54.7\% & 51.0\% & 51.4\%\\
	   ARREST        & 86.6\% & \textbf{57.7}\% & \textbf{53.3}\% & 49.4\% & \textbf{49.3}\% & 90.2\% & \textbf{63.0}\% & \textbf{56.8}\% & \textbf{52.4}\% & \textbf{53.4}\% \\  \hline
  		\end{tabular}
  		}
\end{table*}

\section{Detail of Other Knowledge Distillation Methods}
\label{sec:app_kd_other}

In this appendix, we formulate the other knowledge distillation methods used in Table~\ref{tbl:comp_const}.
We used two methods for our comparisons, (i)~logit and (ii)~attention map.
Hereafter, we formulate them in order.

First, logit~\cite{Cui21} guides the DNN with the final output of the pretrained DNN $\theta_{\rm s}^*$.
It replaces $\mathcal{L}_{\rm RGKD}$ in Eq.~(\ref{formula:3}) to $\mathcal{L}_{\rm logit}$ that is formulated as
\begin{align}
\centering
    \mathcal{L}_{\rm logit} (\bm{x}, \bm{\delta}, \theta_{\rm r})
    = d\left(f(\bm{x} + \bm{\delta}; \theta_{\rm r}),f(\bm{x}; \theta_{\rm s}^*)\right).
\label{formula:app_logit}
\end{align}
We set the hyperparameter $\lambda=1$ in accordance with Cui~\etal~\cite{Cui21}.

Second, attention map~\cite{Wang21} guides the DNN with a spatial attention map computed from the latent representations~\cite{Zagoruyko17}.
It replaces $\mathcal{L}_{\rm RGKD}$ in Eq.~(\ref{formula:3}) to $\mathcal{L}_{\rm AT}$ that is formulated as
\begin{align}
\centering
    \mathcal{L}_{\rm AT} &= d(\mathrm{AT}(h(\bm{x} + \bm{\delta}; \theta_{\rm r}),\mathrm{AT}(h(\bm{x}; \theta_{\rm s}^*))). \\
    \text{Here, }~ \mathrm{AT}(A) &= \frac{F(A)}{||F(A)||_2}, ~F(A) = \sum^C_i|A_i|.
\label{formula:attention}
\end{align}
Note that this loss function assumes that the latent representations $h(\bm{x} + \bm{\delta}; \theta_{\rm r})$ and $h(\bm{x}; \theta_{\rm s}^*)$ are obtained from convolutional network~\cite{Krizhevsky12,Simonyan14c_ICLR,Szegedy14,He16,Zagoruyko16}, and tensors in $\mathds{R}^{C \times H \times W}$.
$A_i$~($\in \mathds{R}^{H \times W}$) indicates the $i$th tensor along $C$~(channel) direction in the representation.
Therefore, in Eq.~(\ref{formula:attention}), $F(A)$ means simply summing up the representation along $C$ direction, and $\mathrm{AT}(A)$ normalizes it.
We set the hyperparameter $\lambda=2$ in accordance with Wang~\etal~\cite{Wang21}.

We used the same experimental settings in AFT and RGKD except for the settings described above.

\section{Detail of Optimization with Additional Examples}
\label{sec:app_add_data}

In this appendix, we explain the setup of our experiments with additional examples~(listed in Table~\ref{tbl:RST}).
In these experiments, we utilized the framework in RST~\cite{Raghunathan20}.
The experimental settings were mostly the same as those explained in Subsection~\ref{subsect:exp_set}.
Thus, here, we describe only the settings that differed from those settings.

In the experiments, we utilized the additional unlabeled examples that were originally provided by Carmon~\etal~\cite{Carmon19}.
These examples were pseudo labeled in advance~\cite{Carmon19}, and we used them as ground-truth labels.
The batch size was set to 256, and the fraction of unlabeled examples in the batch was 0.5, \ie, 128 unlabeled examples were used in each batch.
In RST in Table~\ref{tbl:RST}, we set the number of training epochs to 200.
The learning rate started at 0.1 and then decayed by $\times 0.1$ with transition epochs $\{150, 180\}$.
These settings follow Raghunathan~\etal~\cite{Raghunathan20}.
In ARREST, we set the number of training epochs to 100.
The learning rate started at 0.025, decayed to 0.02 at 50 epochs, and then decayed by half every 10 epochs thereafter.

Finally, in accordance with Raghunathan~\etal~\cite{Raghunathan20}, $\mathcal{L}_{\rm CE}$ is calculated with not only adversarial but also clean examples in both RST and ARREST.
The resulting loss function is denoted as $\mathcal{L}_{\rm CE\_RST}$,
\begin{align}
\centering
    \mathcal{L}_{\rm CE\_RST} (\bm{x}, \bm{\delta}, y, \theta_{\rm r}) =
    0.5\ \mathcal{L}_{\rm CE} (f(\bm{x} + \bm{\delta}; \theta_{\rm r}), y) + 
    0.5\ \mathcal{L}_{\rm CE} (f(\bm{x}; \theta_{\rm r}), y).
\label{formula:app_ce}
\end{align}
We replaced $\mathcal{L}_{\rm CE}$ with $\mathcal{L}_{\rm CE\_RST}$ in optimization.
Note that the PGD objective function was maintaining $\mathcal{L}_{\rm CE}(f(\bm{x} + \bm{\delta}; \theta_{\rm r}), y)$ alone, without $\mathcal{L}_{\rm CE}(f(\bm{x}; \theta_{\rm r}), y)$.

\end{document}